\documentclass{article}

\usepackage{microtype}
\usepackage{graphicx}
\usepackage{subcaption}
\usepackage{booktabs}
\usepackage{xcolor}
\usepackage{url}

\usepackage[preprint]{preprint}

\usepackage{amsmath}
\usepackage{amssymb}
\usepackage{mathtools}
\usepackage{amsthm}
\usepackage{bm}

\usepackage[capitalize,noabbrev]{cleveref}

\usepackage{wrapfig}
\usepackage{multirow}
\usepackage{enumitem}
\usepackage{siunitx}
\newcommand{\ours}{ThinkProprio}

\def\vq{{\bm{q}}}

\def\va{{\bm{a}}}

\DeclareMathOperator{\softmax}{softmax}

\theoremstyle{plain}

\theoremstyle{definition}

\theoremstyle{remark}

\usepackage[disable,textsize=tiny]{todonotes}

\title{Think Proprioceptively: State-Grounded Visual Token Selection for VLA Policies}

\author{
  Fangyuan~Wang\textsuperscript{1,2} \quad
  Peng~Zhou\textsuperscript{3} \quad
  Jiaming~Qi\textsuperscript{4} \quad
  Shipeng~Lyu\textsuperscript{1,2} \\
  Chengyang~He\textsuperscript{5} \quad
  David~Navarro-Alarcon\textsuperscript{1,*} \quad
  Guodong~Guo\textsuperscript{2,*} \\
  \normalfont\textsuperscript{1}The Hong Kong Polytechnic University \quad
  \normalfont\textsuperscript{2}Eastern Institute of Technology \\
  \normalfont\textsuperscript{3}Great Bay University \quad
  \normalfont\textsuperscript{4}Northeast Forestry University \quad
  \normalfont\textsuperscript{5}National University of Singapore
}

\hypersetup{
  pdfauthor={Fangyuan Wang, Peng Zhou, Jiaming Qi, Shipeng Lyu, Chengyang He, Guodong Guo, David Navarro-Alarcon},
  pdftitle={Think Proprioceptively: State-Grounded Visual Token Selection for VLA Policies},
  pdfsubject={arXiv preprint}
}

\begin{document}

\maketitle
\begingroup
\renewcommand{\thefootnote}{*}
\footnotetext{Corresponding authors.}
\endgroup

\begin{center}
  \centering
  \includegraphics[width=\linewidth]{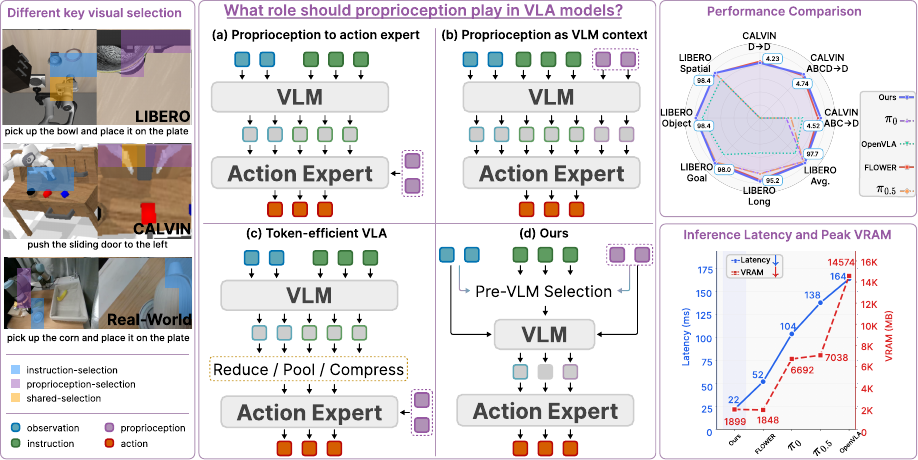}
  \captionof{figure}{\ours{} investigates what role proprioception should play in VLA models.
\textbf{Left:} Across simulation, and real-world tasks, instruction selection (blue) and state selection (purple) highlight complementary visual evidence.
\textbf{Middle:} Four representative VLA designs: (a) proprioception as late input to the action expert, (b) proprioception as VLM context tokens, (c) generic token compression before the action expert, and (d) \ours{} uses both language and proprioception to gate visual tokens \emph{before} VLM. \textbf{Right:} \ours{} matches strong baselines across CALVIN and LIBERO while achieving the lowest inference latency with a compact visual-token budget.
}
  \label{fig:teaser}
\end{center}

\begin{abstract}
  Vision-language-action (VLA) models typically inject proprioception only as a late conditioning signal, preventing robot state from grounding instruction understanding or directing visual attention. We introduce \ours{}, which discretizes proprioception into VLM-vocabulary tokens and uses them jointly with the instruction to gate visual patches before VLM computation, steering the model toward action-relevant evidence while discarding redundant tokens early. We find that proprioception added as a passive conditioning signal leaves performance essentially unchanged; its value emerges when token-form state acts as an active query that, with the instruction, selects which visual patches the VLM processes. Systematic ablations show that VLM-vocabulary tokens outperform learned projectors as the state encoding, and that retaining only about \SI{12}{\percent} of the visual tokens surpasses on CALVIN ABC$\to$D. Across CALVIN, LIBERO, and real-world manipulation, \ours{} reduces end-to-end inference latency while improving the matched full-token baseline.

\end{abstract}

\keywords{Robot manipulation, Vision-language-action models}

\section{Introduction}
\label{sec:intro}
Vision-language-action (VLA) models translate visual observations and language instructions into executable actions through large-scale pretraining \cite{zitkovich2023rt,kim2025openvla,black2024pi0}. Yet contact-rich manipulation
depends not only on what is visible and what is requested, but also on the robot's embodiment, including its joint configuration and motion. Most current VLA pipelines treat proprioception as a late conditioning signal for the action head, couple it only weakly to perception, or omit it altogether. This raises a concrete design question: \emph{should robot state be a late conditioning signal for action generation, or an active participant in instruction grounding and visual attention?}

Holding the backbone, action head, data, and training budget fixed, we vary only how proprioception is encoded and where it enters, finetuning on CALVIN ABC$\to$D (\Cref{tab:ablation-encoding}). Passive late conditioning at the action head matches omitting state entirely, and an MLP projection into the VLM hurts; only VLM-vocabulary tokens routed through the VLM improve over the no-proprioception baseline.

\begin{wraptable}[12]{r}{0.5\linewidth}
  \centering
  \vspace{-1.0\baselineskip}
  \caption{Taxonomy of representative VLA designs. Subheaders specify the order of slash-separated fields. Tok. = tokenized and Comp. = compressed.}
  \label{tab:taxonomy}
  \scriptsize
  \setlength{\tabcolsep}{2pt}
  \renewcommand{\arraystretch}{0.95}
  \resizebox{0.98\linewidth}{!}{%
    \begin{tabular}{@{}lll@{}}
      \toprule
      \textbf{Model} & \textbf{VL} & \textbf{Proprio.} \\
                      & {\scriptsize repr. / act. cond.} & {\scriptsize enc. / entry / act. cond.} \\
      \midrule
      $\pi_0$~\cite{black2024pi0}       & Dense / Cross-attn  & MLP / ACT / Cross-attn \\
      $\pi_{0.5}$~\cite{black2025pi05}  & Dense / Cross-attn  & Tok. / VLM / Cross-attn \\
      SmolVLA~\cite{shukor2025smolvla}  & Dense / Cross-attn  & MLP / VLM / Cross-attn \\
      FLOWER~\cite{reuss2025flower}     & Dense / Cross-attn  & MLP / ACT / AdaLN \\
      Dita~\cite{hou2025dita}           & Dense / In-context  & MLP / ACT / In-context \\
      CogACT~\cite{li2024cogact}        & Comp. / In-context  & N/A \\
      OTTER~\cite{huang2025otter}       & Pooled / Cross-attn & MLP / ACT / Cross-attn \\
      DiT-Blocks~\cite{dasari2024ditpi} & Pooled / AdaLN      & MLP / VLM / Cross-attn \\
      \bottomrule
    \end{tabular}%
  }
\end{wraptable}
In published systems, proprioceptive design is typically entangled with vision-language aggregation and the action conditioning mechanism (see \Cref{tab:taxonomy}), making the functional role of proprioception hard to attribute. We therefore vary these axes systematically. Prior work treats efficient visual-token selection~\cite{jiang2025lightvla,rao2021dynamicvit} and routing proprioception through the VLM token interface~\cite{black2025pi05} as separate developments; once proprioception lives in this token space, it can also serve as a query signal for visual evidence, complementary to the instruction.

We instantiate this idea as \ours{}, which gates visual patches before VLM computation using separate instruction and proprioception branches. Our contributions are:
\begin{itemize}[leftmargin=*,itemsep=2pt,topsep=2pt]
  \item A systematic study isolating proprioceptive encoding and entry point from vision-language aggregation and action conditioning, showing that encoding proprioception as VLM-vocabulary tokens both preserves baseline performance and exposes state to downstream visual reasoning.
  \item \ours{}, a state-grounded visual-token gating mechanism that uses language and proprioception as complementary query branches before VLM computation.
  \item Empirical evaluation on CALVIN, LIBERO, and a 14-task real-world benchmark, where \ours{} matches or surpasses strong baselines at a fraction of the visual tokens and the lowest per-step latency.
\end{itemize}

\section{Related Work}
\label{sec:related-work}

\noindent\textbf{Vision-Language Feature Extraction.}
VLA systems differ in how vision-language backbone outputs are presented to the action head. Some pass dense token sets directly, as in $\pi_0$~\cite{black2024pi0}, $\pi_{0.5}$~\cite{black2025pi05}, and FLOWER~\cite{reuss2025flower}, preserving fine-grained information but increasing computation. Others aggregate tokens before control: CogACT~\cite{li2024cogact} compresses the set, while DiT-Block~\cite{dasari2024ditpi}, OTTER~\cite{huang2025otter}, and LightVLA~\cite{jiang2025lightvla} apply pooling-style reductions that inherit from the broader token-reduction literature~\cite{rao2021dynamicvit,ryoo2021tokenlearner}. These methods span a spectrum from indiscriminate aggregation, which reduces tokens without regard to task content, to guided selection that conditions retention on an external signal.

\noindent\textbf{Proprioceptive Encoding and Entry.} Prior work integrates proprioception with different encodings and entry points. $\pi_{0.5}$~\cite{black2025pi05} serializes proprioception as text before VLM input, whereas \ours{} maps discretized state bins directly to VLM vocabulary IDs. GR00T-N1~\cite{nvidia2024groot} and SmolVLA~\cite{shukor2025smolvla} instead project proprioception with multilayer perceptrons into the VLM feature space, which increases representational flexibility but can introduce mismatch relative to pretrained token features. FLOWER~\cite{reuss2025flower} conditions the action head directly on proprioceptive inputs, bypassing the backbone, and some single-system VLAs such as CogACT~\cite{li2024cogact} omit explicit proprioceptive input entirely. Encoding and entry point co-vary in the literature: MLP-encoded state is almost always fed to the action head, token-form state enters the VLM, and proprio-free designs cluster among single-system policies. This coupling has historically been treated as a fixed architectural package rather than as two independent design axes.

\noindent\textbf{Action Conditioning Mechanisms.}
VLA action heads adopt three main conditioning mechanisms. \emph{In-context} approaches concatenate vision-language and proprioceptive embeddings with the action sequence, as in Dita~\cite{hou2025dita} and decoder-only systems such as OpenVLA~\cite{kim2025openvla}. \emph{AdaLN-style} modulation predicts layer-wise scale and shift from pooled features and is used by DiT-Block~\cite{dasari2024ditpi}, FLOWER~\cite{reuss2025flower}, and MDT~\cite{reuss2024mdt}. \emph{Cross-attention} retains token-level access at higher cost and is used by $\pi_0$~\cite{black2024pi0}, $\pi_{0.5}$~\cite{black2025pi05}, and FLOWER's Flow Transformer~\cite{reuss2025flower}. The three families trade expressivity for cost: cross-attention is most expressive but scales with the context-token count, AdaLN is cheapest but collapses tokens into a global modulation, and in-context conditioning falls in between.

Across these methods, proprioception is treated as a passive conditioning signal, but not used to select what the policy looks at. \ours{} departs from this convention by using state as an active query against visual tokens, alongside the instruction, before the VLM consumes them.

\section{Method}
\label{sec:method}
\begin{figure*}[!t]
  \centering
  \includegraphics[width=1.0\textwidth]{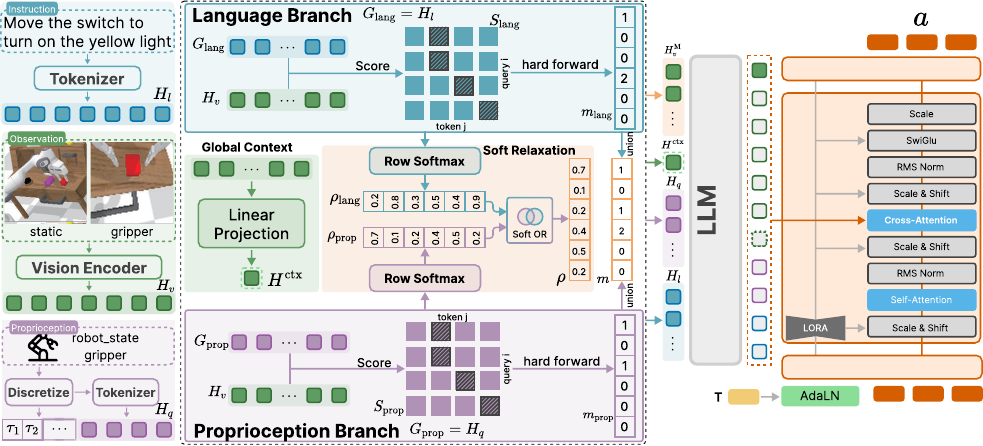}
  \caption{Overview of \ours{}. Proprioception is used with the instruction in two guidance branches to select task-relevant visual patches before VLM computation, alongside a global context token. The action head attends to the resulting features to generate actions.}
  \label{fig:overview}
\end{figure*}

We consider a policy $\pi_\theta$ composed of a vision-language backbone and a separate action head. At timestep~$t$, the policy receives an observation $o_t$ comprising $n$ RGB images $(I_t^1,\ldots,I_t^n)$, a language instruction $\ell$, and a proprioceptive state $\vq_t$ that encodes the robot's current configuration, including joint angles and end-effector pose. We represent each input stream as a token sequence with shared embedding dimension $D$: vision tokens $H_v \in \mathbb{R}^{N_v \times D}$, language tokens $H_l \in \mathbb{R}^{N_l \times D}$, and proprioceptive tokens $H_q \in \mathbb{R}^{N_q \times D}$, where $N_v$, $N_l$, and $N_q$ denote the corresponding token counts. As shown in \cref{fig:overview}, \ours{} instantiates this dual-system policy by discretizing proprioception into VLM-vocabulary tokens and using instruction and proprioceptive tokens in two separate guidance branches to select visual patches before VLM computation. The backbone $f_{\text{VLM}}$ maps the resulting compact multimodal sequence to conditioning features $C$, and the action head $f_{\text{ACT}}$ predicts a continuous action chunk $\va_{t:t+\mathcal{H}}$ conditioned on $C$.

\subsection{Proprioceptive State Encoding}
\label{sec:method-encoding}
The proprioceptive state $\vq_t$ contains $d_q$ scalar values, so $N_q=d_q$. We discretize each value with uniform binning over a clipped range, where $q_{\min}$ and $q_{\max}$ are shared scalar clipping bounds applied to each dimension and $B$ is the number of bins. For each state element $q_{t,k}$ with $k \in \{1,\dots,d_q\}$, we compute the bin index
\begin{equation}
  b_{t,k} = \left\lfloor \frac{\operatorname{clip}(q_{t,k}, q_{\min}, q_{\max}) - q_{\min}}{q_{\max} - q_{\min}} \cdot (B-1) \right\rfloor.
\end{equation}

We map each bin index to a proprioceptive token ID using the reverse mapping $\tau_{t,k} = V - 1 - b_{t,k}$, where $V$ is the VLM vocabulary size. These IDs are vocabulary aliases for discretized state values, not the output of the language tokenizer. Reusing the existing embedding table adds no new embedding parameters and uses the same lookup pathway as text inputs; the reverse indexing keeps the proprioceptive token range disjoint from the tokens produced by the language tokenizer for our prompts, so the two streams do not collide. We then obtain the corresponding embeddings from the VLM token embedding table, $H_q = \operatorname{Embed}(\bm{\tau}_t)$, where $\bm{\tau}_t = [\tau_{t,1},\dots,\tau_{t,d_q}]$.

\subsection{Embodied Visual Token Gating}
\label{sec:method-integration}
We gate visual tokens before they enter the VLM so that subsequent VLM computation operates on a compact set of action-relevant patches. The selector uses two complementary guidance branches, one driven by the instruction and one by proprioception. The instruction branch emphasizes task-semantic evidence such as objects and goals, while the proprioception branch emphasizes configuration-dependent evidence such as the gripper and contact regions.

\noindent\textbf{Branch-specific scoring.}
Let $\Omega=\{\mathrm{lang},\mathrm{prop}\}$ index the two branches, with guidance tokens $G_{\mathrm{lang}}=H_l$ and $G_{\mathrm{prop}}=H_q$. We normalize the visual and guidance tokens as $\tilde H_v=\mathrm{RMSNorm}(H_v)$ and $\tilde G_\omega=\mathrm{RMSNorm}(G_\omega)$ for each $\omega\in\Omega$. Rather than scoring visual tokens directly from the guidance tokens, we use a vote-based construction that conditions each vote on a visual token, so the selector forms visual-context-dependent votes instead of a single top-down saliency map. For branch $\omega$, we compute
\begin{equation}
  Q_\omega =
  \softmax\left(\frac{\tilde H_v \tilde G_\omega^\top}{\sqrt{D}}\right)\tilde G_\omega ,
  \qquad
  S_\omega =
  \frac{\mathrm{RMSNorm}(Q_\omega)\tilde H_v^\top}{\sqrt{D}} .
\end{equation}
Each row of $Q_\omega$ is obtained by letting visual token $i$ attend to the branch guidance tokens and extract the guidance information most relevant to it. The resulting guidance-conditioned query then scores all visual tokens as candidates for retention. The matrix $S_\omega\in\mathbb{R}^{N_v\times N_v}$ is therefore a branch-specific vote matrix in which row index $i$ denotes the visual token that issues the query and column index $j$ denotes a candidate visual token for retention; entry $S_\omega[i,j]$ measures how strongly guidance-conditioned visual token $i$ votes for retaining visual token $j$.

\noindent\textbf{Vote-based gating with straight-through relaxation.}
Each branch converts its score matrix into a hard selection mask for the forward pass and a soft relaxation for gradient propagation. During training, we perturb each branch score matrix as $\hat S_\omega=S_\omega+\alpha\Gamma_\omega$, where $\Gamma_\omega$ is sampled elementwise from the standard Gumbel distribution and $\alpha$ is cosine-annealed from $\alpha_{\mathrm{start}}$ to $\alpha_{\mathrm{end}}$. This perturbation encourages exploration early in training and approaches deterministic selection near convergence. At inference time, we remove the perturbation by setting $\hat S_\omega=S_\omega$.

In the hard forward path, each row $i$ casts one vote for a candidate visual token, $z_\omega[i] = \arg\max_{j\in\{1,\ldots,N_v\}}\hat S_\omega[i,j]$. Branch $\omega$ selects token $j$ if it receives at least one vote, $m_\omega[j]=\mathbf{1}[\exists\, i:\, z_\omega[i]=j]$, and we merge the two branch masks by union,
\begin{equation}
  m[j] = m_{\mathrm{lang}}[j] \vee m_{\mathrm{prop}}[j],
  \qquad
  M=\{j:m[j]=1\}.
\end{equation}

In parallel, we compute the soft relaxation used for backpropagation. The row-wise probabilities $P_\omega = \softmax(\hat S_\omega)$ give $P_\omega[i,j]$ as the relaxed probability that row $i$ votes for token $j$. Treating per-row votes as conditionally independent, the probability that token $j$ receives at least one vote within branch $\omega$ is the \emph{noisy-OR}~\cite{pearl1988probabilistic} over rows (see Appendix \Cref{app:selector-preliminaries}), $\rho_\omega[j] = 1 - \prod_{i=1}^{N_v}\!\left(1 - P_\omega[i,j]\right)$. We aggregate the two branches by an analogous noisy-OR over branches:
\begin{equation}
  \rho[j] = 1 - \bigl(1 - \rho_{\mathrm{lang}}[j]\bigr)\bigl(1 - \rho_{\mathrm{prop}}[j]\bigr).
\end{equation}

The straight-through gate combines the hard union mask and the soft union probability $w[j] = m[j] + \rho[j] - \mathrm{sg}(\rho[j])$, where $\mathrm{sg}(\cdot)$ denotes stop-gradient. Thus, $w[j]$ equals the hard mask $m[j]$ in the forward pass, while gradients through $w[j]$ follow the soft probability $\rho[j]$. During training, each retained token $H_v[j]$ with $j\in M$ is multiplied by $w[j]$ before being packed into $H_v^M$. At inference time, we use the hard mask directly.

\noindent\textbf{Diversity regularization.}
To prevent the instruction and proprioception branches from selecting identical token sets, we aggregate the unperturbed scores over voting rows, $r_{\omega}[j]=\sum_{i=1}^{N_v} S_{\omega}[i,j]$, and penalize agreement between the resulting normalized selection distributions:
\begin{equation}
  \mathcal{L}_{\mathrm{div}}
  =
  \big[
  \operatorname{sim}_{\cos}\!\big(\softmax(r_{\mathrm{lang}}),\, \softmax(r_{\mathrm{prop}})\big)
  - \gamma
  \big]_+^{\,2},
\end{equation}
where $\operatorname{sim}_{\cos}(\cdot,\cdot)$ denotes cosine similarity and $\gamma$ is the diversity margin. This encourages complementary instruction and proprioception guidance while still allowing the union mask $M$ to retain evidence supported by both branches.

\noindent\textbf{Global context token.}
Aggressive token selection can remove scene-level information that receives little support from individual votes but remains useful for action generation. To preserve coarse visual context, we append a learned global token computed from the full pre-selection visual sequence,
\begin{equation}
  H^{\mathrm{ctx}}=W_c\left(\frac{1}{N_v}\sum_{j=1}^{N_v}H_v[j]\right),
\end{equation}
where $W_c$ is a learned linear projection.

\subsection{Training Objective}
\label{sec:method-objective}
After visual token gating, the VLM maps the compact multimodal sequence to conditioning features, $C=f_{\text{VLM}}([H_v^M;H^{\mathrm{ctx}};H_q;H_l])$. The selector, VLM backbone, and action head are trained end-to-end. We train the action head with flow matching~\cite{lipman2023flowmatching}. We sample a continuous flow time $T \sim \mathrm{Unif}(0,1)$ and construct a noisy action chunk $\va_{t:t+\mathcal{H}}^{T} = (1-T)\va_{t:t+\mathcal{H}} + T\epsilon$, where $\epsilon \sim \mathcal{N}(0, \mathbf{I})$. Following FLOWER, we embed $T$ and inject it into the action Transformer through global AdaLN-style modulation, while the action tokens cross-attend to $C$. Under this linear interpolation, the target velocity field is $\frac{d}{dT}\va_{t:t+\mathcal{H}}^{T}=\epsilon - \va_{t:t+\mathcal{H}}$. The action head predicts $v_\theta = f_{\text{ACT}}(\va_{t:t+\mathcal{H}}^{T}, T, C)$, and we optimize $\mathcal{L}_{\mathrm{fm}} = \mathbb{E}_{T, \epsilon}\!\left[\left\|v_\theta - (\epsilon - \va_{t:t+\mathcal{H}})\right\|^2\right]$. The final training objective combines the flow matching loss with the branch diversity regularizer,
\begin{equation}
  \mathcal{L} = \mathcal{L}_{\mathrm{fm}} + \lambda_{\mathrm{div}}\,\mathcal{L}_{\mathrm{div}}.
\end{equation}
where $\lambda_{\mathrm{div}}$ controls the strength of the diversity regularizer.

\section{Experiments}
\label{sec:experiments}
We address the following research questions: \textbf{RQ1}: How should proprioception be represented and incorporated into a VLA policy? \textbf{RQ2}: Can proprioception guide visual-token selection to retain task- and state-relevant evidence while reducing inference cost? \textbf{RQ3}: Do the resulting design choices improve manipulation performance and efficiency across simulation and real-world tasks?

\subsection{Experiments Setup}
\label{sec:exp-setup}

\noindent\textbf{Baselines.}
On CALVIN, we compare against single-system VLAs such as OpenVLA~\cite{kim2025openvla}, which represents actions as discrete tokens, and methods with dedicated continuous-control action heads, including GR-1~\cite{wu2024gr1}, RoboFlamingo~\cite{li2023roboflamingo}, $\pi_0$~\cite{black2024pi0}, $\pi_{0.5}$~\cite{black2025pi05}, and FLOWER~\cite{reuss2025flower}. We report $\pi_0$ and $\pi_{0.5}$ results from our own fine-tuning runs, marked with $^*$. We further compare with visual planning methods, including SuSIE~\cite{black2024susie}, VPP~\cite{hu2024vpp}, and Seer~\cite{tian2024predictive}. On LIBERO, we compare with strong OpenVLA variants and recent VLA baselines, including OpenVLA-OFT~\cite{kim2025oft}, COA-VLA~\cite{li2025coavla}, LightVLA~\cite{jiang2025lightvla}, WorldVLA~\cite{cen2025worldvla} and SmolVLA~\cite{shukor2025smolvla}. Unless otherwise noted, all reported results are means over 5 seeds.

\begin{wrapfigure}[12]{r}{0.36\linewidth}
  \centering
  \includegraphics[width=\linewidth]{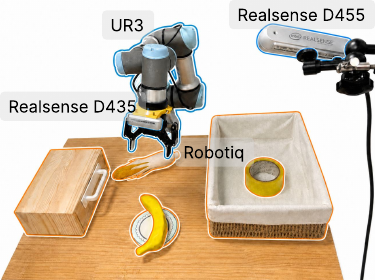}
  \caption{Real-world task setup.}
  \label{fig:real-world}
\end{wrapfigure}
\noindent\textbf{Simulation \& Real-world setup.}
We evaluate \ours{} in both simulation and the real world. \textbf{CALVIN}~\cite{mees2022calvin} evaluates long-horizon control by requiring policies to complete chains of five tasks. \textbf{LIBERO}~\cite{liu2023libero} evaluates generalization across four suites, Spatial, Object, Goal, and Long. For real-world evaluation, we use a UR3 arm with a parallel gripper and two RGB cameras: a fixed third-person view and a wrist-mounted view, as shown in Figure~\ref{fig:real-world}. We collect approximately 120 minutes of teleoperated demonstrations and fine-tune from a pretrained checkpoint. The real-world experiments contains 14 tasks: 12 pick-and-place tasks across three objects (banana, corn, and tape), and two drawer tasks (open and close). The Gumbel exploration noise annealed from 1.0 to 0.01 and a diversity regularizer using weight $5\times10^{-4}$, margin $\gamma=0.75$. Full implementation details are in Appendix \Cref{app:exp-details}.

\subsection{Performance and Efficiency}
\label{sec:exp-main-results}

\noindent\textbf{Simulation performance.}
On CALVIN ABC$\to$D (\Cref{tab:main-sim}a), success drops with chain length, consistent with compounding errors. \ours{} achieves the best Avg.\ Len.\ of \num{4.52}, exceeding FLOWER's \num{4.44}. FLOWER is stronger at LH-1 and LH-2, while \ours{} leads from LH-3 to LH-5, consistent with state-conditioned gating helping most when configuration drift accumulates over longer chains. The gains are strongest at longer horizons: \ours{} reaches \SI{86.4}{\percent} at LH-4 and \SI{79.1}{\percent} at LH-5, reducing FLOWER's LH-5 failure rate by roughly \SI{15}{\percent}. This supports the intuition that state- and instruction-guided token selection helps retain key visual evidence as the configuration evolves. On LIBERO (\Cref{tab:main-sim}b), \ours{} achieves the best overall success, ties on LIBERO-Spatial, and leads on LIBERO-Object and LIBERO-Long.

\begin{table*}[t]
  \centering
  \caption{Simulation benchmark results. CALVIN ABC$\to$D reports success rate (\%) at each chain length and average completed chain length. LIBERO reports success rate (\%) across 10 tasks per suite. $^*$Finetuned by us. Best results are in \textbf{bold}; second-best are \underline{underlined}.}
  \label{tab:main-sim}
  \scriptsize
  \begin{minipage}[t]{0.535\textwidth}
    \centering
    \textbf{(a) CALVIN ABC$\to$D}

    \vspace{0.3em}
    \setlength{\tabcolsep}{2.5pt}
    \resizebox{\linewidth}{!}{%
      \begin{tabular}{@{}lcccccc@{}}
        \toprule
        \textbf{Method} & \textbf{LH-1} $\uparrow$ & \textbf{LH-2} $\uparrow$ & \textbf{LH-3} $\uparrow$ & \textbf{LH-4} $\uparrow$ & \textbf{LH-5} $\uparrow$ & \textbf{Avg.} $\uparrow$ \\
        \midrule
        OpenVLA           & 91.3 & 77.8 & 62.0 & 52.1 & 43.5 & 3.27 \\
        GR-1              & 85.4 & 71.2 & 59.6 & 49.7 & 40.1 & 3.06 \\
        RoboFlamingo      & 82.4 & 61.9 & 46.6 & 33.1 & 23.5 & 2.47 \\
        $\pi_0^{*}$       & 70.0 & 48.0 & 37.0 & 28.0 & 18.0 & 2.01 \\
        $\pi_{0.5}^{*}$   & 71.0 & 56.0 & 45.0 & 37.0 & 29.0 & 2.38 \\
        SuSIE             & 87.0 & 69.0 & 49.0 & 38.0 & 26.0 & 2.69 \\
        VPP               & 95.7 & 91.2 & 86.3 & 81.0 & 75.0 & 4.29 \\
        Seer              & 96.3 & 91.6 & 86.1 & 80.3 & 74.0 & 4.29 \\
        FLOWER            & \textbf{99.3} & \textbf{96.0} & \underline{90.3} & \underline{82.3} & \underline{75.5} & \underline{4.44} \\
        \midrule
        \textbf{\ours{}}  & \underline{98.9} & \underline{95.4} & \textbf{91.6} & \textbf{86.4} & \textbf{79.1} & \textbf{4.52} \\
        \bottomrule
      \end{tabular}
    }
  \end{minipage}
  \hfill
  \begin{minipage}[t]{0.455\textwidth}
    \centering
    \textbf{(b) LIBERO}

    \vspace{0.3em}
    \setlength{\tabcolsep}{2.5pt}
    \resizebox{\linewidth}{!}{%
      \begin{tabular}{@{}lccccc@{}}
        \toprule
        \textbf{Method} & \textbf{Spa.} $\uparrow$ & \textbf{Obj.} $\uparrow$ & \textbf{Goal} $\uparrow$ & \textbf{Long} $\uparrow$ & \textbf{Avg.} $\uparrow$ \\
        \midrule
        OpenVLA           & 84.7 & 88.4 & 79.2 & 53.7 & 76.5 \\
        WorldVLA          & 85.6 & 89.0 & 82.6 & 59.0 & 79.1 \\
        SmolVLA           & 93.0 & 94.0 & 91.0 & 77.0 & 88.8 \\
        OpenVLA-OFT       & 97.6 & 98.4 & 97.9 & 94.5 & 97.1 \\
        COA-VLA           & 85.3 & 93.1 & 85.8 & 55.0 & 79.8 \\
        $\pi_0$           & 96.8 & 98.8 & 95.8 & 85.2 & 94.2 \\
        $\pi_{0.5}$       & \underline{98.0} & 97.8 & 95.6 & 85.8 & 94.3 \\
        FLOWER            & 97.5 & \underline{99.1} & 96.1 & \underline{94.9} & 96.9 \\
        LightVLA          & \textbf{98.4} & 98.4 & \textbf{98.2} & 94.6 & \underline{97.4} \\
        \midrule
        \textbf{\ours{}}  & \textbf{98.4} & \textbf{99.2} & \underline{98.0} & \textbf{95.2} & \textbf{97.7} \\
        \bottomrule
      \end{tabular}
    }
  \end{minipage}
\end{table*}

\begin{table}[t]
  \centering
  \caption{Real-world task success rates. Left: pick-place tasks. Right: drawer tasks.}
  \label{tab:realworld-success}
  \small
  \setlength{\tabcolsep}{4pt}
  \resizebox{\linewidth}{!}{%
    \begin{tabular}{@{}lcccccc@{\hspace{1em}}c@{\hspace{1em}}lcc@{}}
      \toprule
      \textbf{Pick-place} & \multicolumn{2}{c}{\textbf{Banana}} & \multicolumn{2}{c}{\textbf{Corn}} & \multicolumn{2}{c}{\textbf{Tape}} & & & \multicolumn{2}{c}{\textbf{Drawer}} \\
      \cmidrule(lr){2-3} \cmidrule(lr){4-5} \cmidrule(lr){6-7} \cmidrule(l){10-11}
      & \textbf{FLOWER} & \textbf{\ours{}} & \textbf{FLOWER} & \textbf{\ours{}} & \textbf{FLOWER} & \textbf{\ours{}} & & & \textbf{FLOWER} & \textbf{\ours{}} \\
      \midrule
      Basket $\rightarrow$ Plate  & 16/20 & 17/20 & 17/20 & 19/20 & 16/20 & 19/20 & \multirow{4}{*}{\rule{0.35pt}{3.6em}} & \multirow{2}{*}{Open}  & \multirow{2}{*}{16/20} & \multirow{2}{*}{18/20} \\
      Basket $\rightarrow$ Table  & 16/20 & 18/20 & 15/20 & 18/20 & 17/20 & 19/20 & & & & \\
      Plate $\rightarrow$ Basket  & 19/20 & 17/20 & 15/20 & 16/20 & 16/20 & 16/20 & & \multirow{2}{*}{Close} & \multirow{2}{*}{16/20} & \multirow{2}{*}{18/20} \\
      Table $\rightarrow$ Basket  & 16/20 & 18/20 & 15/20 & 17/20 & 16/20 & 19/20 & & & & \\
      \bottomrule
    \end{tabular}
  }
\end{table}

\noindent\textbf{Real-world robot evaluation.}
We compare against FLOWER as the strongest baseline for real world experiments. \Cref{tab:realworld-success} reports successful trials out of 20 for each task. \ours{} achieves an overall success rate of \SI{88.9}{\percent} against \SI{80.7}{\percent} for FLOWER. Gains are consistent across objects and task categories: \ours{} matches or exceeds FLOWER on 13 of 14 tasks, with the single exception being Banana Plate$\rightarrow$Basket (17/20 vs.\ 19/20). Drawer manipulation shows a uniform improvement of 18/20 versus 16/20 on both open and close. These results confirm that selected visual tokens provide consistent gains across diverse task configurations.

\begin{wraptable}[9]{r}{0.42\textwidth}
  \centering
  \vspace{-0.8\baselineskip}
  \caption{Inference efficiency on CALVIN ABC$\to$D. Latency is per timestep in ms; VRAM is MB.}
  \label{tab:efficiency}
  \small
  \setlength{\tabcolsep}{3.5pt}
  \resizebox{\linewidth}{!}{%
    \begin{tabular}{@{}lcccc@{}}
      \toprule
      \textbf{Method} & \textbf{Tokens} & \textbf{Latency} & \textbf{VRAM} & \textbf{Avg. Len.} \\
      \midrule
      OpenVLA           & 256 & 164 & 14574 & 3.27 \\
      $\pi_0$           & 256 & 104 & 6692 & 2.01  \\
      $\pi_{0.5}$       & 256 & 138 & 7038 & 2.38  \\
      FLOWER            & 100 & 52  & 1848 & 4.44  \\
      \midrule
      \textbf{\ours{}}  & \bfseries 12 & \bfseries 22 & \bfseries 1899 & \bfseries 4.52  \\
      \bottomrule
    \end{tabular}
  }
\end{wraptable}
\noindent\textbf{Computational efficiency.}
\Cref{tab:efficiency} compares inference cost on CALVIN ABC$\to$D. \ours{} keeps only about \num{12} of \num{100} visual tokens per step on average, shortening the sequence processed by both the VLM and the action head. The selector gathers the kept tokens, pads to the per-batch max, and supplies an attention mask. Despite this overhead, \ours{} reaches lower end-to-end latency than FLOWER (\SI{22}{\milli\second} vs.\ \SI{52}{\milli\second}). Peak VRAM grows marginally over FLOWER from selector parameters, still far below OpenVLA.

\subsection{Token Selection Analysis}

\begin{figure}[t]
  \centering
  \includegraphics[width=\linewidth]{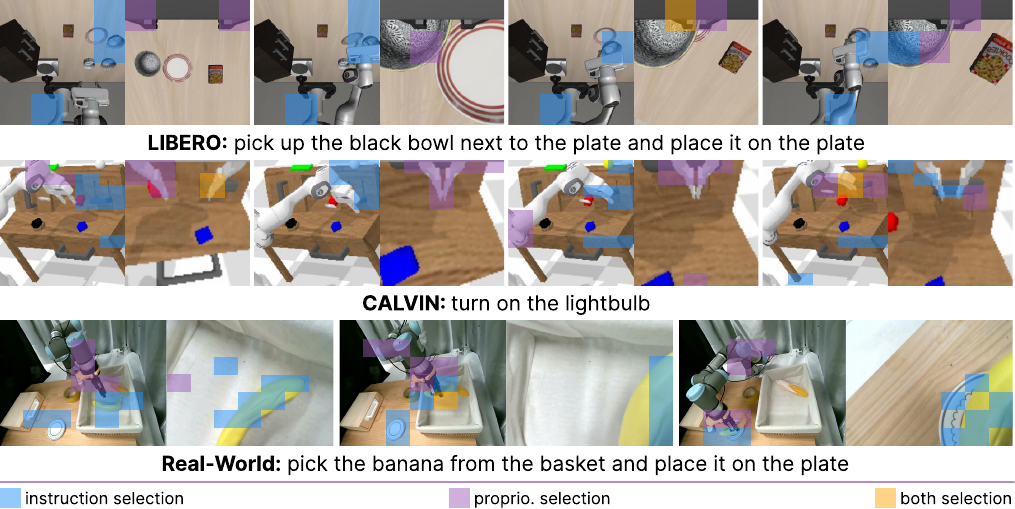}
  \caption{Qualitative token-selection results on LIBERO, CALVIN, and the real-world setup. Across time, the retained tokens track task-relevant objects, targets, and robot--object interaction regions.}
  \label{fig:token-selection}
\end{figure}

\noindent\textbf{Qualitative behavior.} \Cref{fig:token-selection} visualizes retained tokens across simulation and real-world rollouts. The two streams play complementary roles: instruction selected tokens cover task-referenced objects and target regions, while proprioception selected tokens track the gripper and its immediate contact area, with overlap concentrated at the moment of interaction.

\begin{wraptable}[8]{r}{0.45\textwidth}
  \vspace{-0.8\baselineskip}
  \centering
  \caption{Retained-token counts by query source. Both gives their overlap.}
  \label{tab:token-source-breakdown}
  \scriptsize
  \setlength{\tabcolsep}{3pt}
  \renewcommand{\arraystretch}{0.92}
  \resizebox{\linewidth}{!}{%
    \begin{tabular}{@{}lccc@{}}
      \toprule
      \textbf{Benchmark} & $\mathrm{lang}$ & $\mathrm{prop}$ & \textbf{Both} \\
      \midrule
      CALVIN & $8.60{\pm}0.36$ & $3.52{\pm}0.39$ & $0.25{\pm}0.10$ \\
      LIBERO-Spa. & $4.78{\pm}0.83$ & $3.37{\pm}0.57$ & $0.35{\pm}0.08$ \\
      LIBERO-Obj. & $4.48{\pm}0.44$ & $2.58{\pm}0.27$ & $0.18{\pm}0.14$ \\
      LIBERO-Goal & $3.43{\pm}0.48$ & $2.60{\pm}0.50$ & $0.10{\pm}0.08$ \\
      LIBERO-Long & $4.17{\pm}0.49$ & $3.47{\pm}0.17$ & $0.17{\pm}0.08$ \\
      Real-World & $9.36{\pm}0.42$ & $5.87{\pm}0.28$ & $0.61{\pm}0.13$ \\
      \bottomrule
    \end{tabular}
  }
  \vspace{-0.8em}
\end{wraptable}

\noindent\textbf{Complementary query sources.} \Cref{tab:token-source-breakdown} decomposes retained tokens by query source. The overlap stays below one token across all benchmarks, an order of magnitude smaller than either stream alone, so the proprioception branch contributes genuinely complementary evidence rather than re-selecting language-grounded objects. The retained set thus extends beyond a language-conditioned object mask to include configuration-dependent interaction cues.

\subsection{Ablation Studies}
\label{sec:exp-ablation}
We isolate the design choices behind \ours{} with others fixed, ablating each component's contribution and the entry point for proprioception driving visual-token selection.

\begin{wraptable}[15]{r}{0.56\textwidth}
  \centering
  \vspace{-0.8\baselineskip}
  \caption{Ablations on CALVIN ABC$\to$D and LIBERO-Spatial. Tokens denotes the retained visual-token budget.}
  \label{tab:ablation-components}
  \scriptsize
  \setlength{\tabcolsep}{2pt}
  \renewcommand{\arraystretch}{0.92}
  \resizebox{\linewidth}{!}{%
    \begin{tabular}{@{}lcccc@{}}
      \toprule
      & \multicolumn{2}{c}{\textbf{CALVIN ABC$\to$D}} & \multicolumn{2}{c}{\textbf{LIBERO-Spatial}} \\
      \cmidrule(lr){2-3} \cmidrule(l){4-5}
      \textbf{Method variant} & \textbf{Tokens} & \textbf{Avg.\ Len.} $\uparrow$ & \textbf{Tokens} & \textbf{SR} $\uparrow$ \\
      \midrule
      FLOWER & All & $4.44 {\pm} 0.03$ & All & $97.5 {\pm} 0.08$ \\
      +Proprio tokens & All & $4.45 {\pm} 0.02$ & All & $97.8 {\pm} 0.05$ \\
      +Token selection & $\sim$\SI{12}{\percent} & $4.35 {\pm} 0.04$ & $\sim$\SI{23}{\percent} & $97.2 {\pm} 0.06$ \\
      +$H^{\mathrm{ctx}}$ & $\sim$\SI{12}{\percent} & $4.48 {\pm} 0.03$ & $\sim$\SI{23}{\percent} & $98.2 {\pm} 0.04$ \\
      +Div. loss & $\sim$\SI{12}{\percent} & $\mathbf{4.52 {\pm} 0.02}$ & $\sim$\SI{23}{\percent} & $\mathbf{98.4 {\pm} 0.03}$ \\
      \midrule
      $\Omega=\{\mathrm{lang}\}$ & $\sim$\SI{8}{\percent} & $3.40 {\pm} 0.06$ & $\sim$\SI{14}{\percent} & $96.6 {\pm} 0.05$ \\
      $\Omega=\{\mathrm{prop}\}$ & $\sim$\SI{3}{\percent} & $3.12 {\pm} 0.10$ & $\sim$\SI{10}{\percent} & $96.0 {\pm} 0.04$ \\
      \midrule
      Mean pooling & \SI{1}{\percent} & $4.20 {\pm} 0.02$ & \SI{1}{\percent} & $90.3 {\pm} 0.04$ \\
      Max pooling & \SI{25}{\percent} & $4.33 {\pm} 0.03$ & \SI{25}{\percent} & $92.8 {\pm} 0.02$ \\
      Random & \SI{50}{\percent} & $4.24 {\pm} 0.02$ & \SI{50}{\percent} & $87.4 {\pm} 0.08 $ \\
      \bottomrule
    \end{tabular}
  }
  \vspace{-0.8em}
\end{wraptable}
\noindent\textbf{Component ablation.}
In \Cref{tab:ablation-components}, VLM-vocabulary proprioceptive tokens give a small but consistent gain across CALVIN and LIBERO-Spatial. Visual token selection alone reduces the visual-token budget but slightly hurts performance, suggesting that local selection can discard scene-level evidence needed for action generation. Adding the global context token $H^{\mathrm{ctx}}$ recovers this loss, and the diversity loss gives the best result on both benchmarks by encouraging selected tokens to cover complementary evidence.

\noindent\textbf{Query signal for token retention.}
The lower panel of \Cref{tab:ablation-components} compares practical retention strategies. Task-agnostic pooling and random baselines retain much of FLOWER's performance, indicating substantial visual redundancy in CALVIN. Query-guided retention is sensitive to the guidance signal: instruction-only captures task-referenced objects but lacks embodiment context, proprioception-only is too narrow without task semantics, and only their combination keeps both object-centric and configuration-dependent cues. The \num{1.1} Avg.\ Len.\ gap on CALVIN exceeds any plausible budget effect, and the near-zero cross-branch overlap indicates the streams are complementary rather than redundant.

\begin{wraptable}[8]{r}{0.42\textwidth}
  \centering
  \vspace{-0.8\baselineskip}
  \caption{Proprioceptive integration ablation on CALVIN ABC$\to$D.}
  \label{tab:ablation-encoding}
  \small
  \setlength{\tabcolsep}{4pt}
  \resizebox{\linewidth}{!}{%
    \begin{tabular}{@{}lllc@{}}
      \toprule
      \textbf{Encoding} & \textbf{Entry} & \textbf{ACT cond.} & \textbf{Avg.\ Len.} $\uparrow$ \\
      \midrule
      None & --  & --         & 4.44 \\
      MLP  & ACT & AdaLN       & 4.44 \\
      MLP  & VLM & Cross-attn   & 4.15 \\
      VLM-vocab & VLM & Cross-attn   & \textbf{4.48} \\
      \bottomrule
    \end{tabular}
  }
  \vspace{-0.8em}
\end{wraptable}
\noindent\textbf{Proprioceptive encoding and entry point.}
\Cref{tab:ablation-encoding} isolates proprioceptive integration methods. Where state enters the policy matters as much as whether it is present: late action-head modulation leaves the baseline unchanged, while projecting continuous state into the VLM input hurts performance. VLM-vocabulary proprioceptive tokens avoid this projection mismatch by using the backbone's native embedding lookup rather than a learned feature projector. The standalone gain is modest, but this is the only entry point that both preserves performance and exposes proprioceptive tokens to the downstream visual selector.

\section{Conclusion}
\label{sec:conclusion}
We presented \ours{}, a VLA policy that exposes proprioception as VLM-vocabulary tokens and pairs it with language to guide visual-token retention. Across simulation and real-world experiments, \ours{} improves strong baselines while substantially reducing inference latency. The ablations show that proprioception is most effective when used as an active signal for visual selection.

\section*{Limitations}
\label{sec:limitations}
Our real-world evaluation is confined to a setup with rigid objects, broader embodiments, object categories, and baselines remaining future work. The pre-VLM selector forms a vote matrix, incurring $O(N_v^2 D)$ cost that may bottleneck higher visual resolutions. Finally, although the selector retains human-interpretable evidence, it occasionally preserves uninformative patches such as background, as it is supervised by the action objective rather than explicit relevance signals; tightening selection toward interpretable evidence is a promising direction for future work.

\bibliography{proprio}

@book{pearl1988probabilistic,
  title={Probabilistic reasoning in intelligent systems: networks of plausible inference},
  author={Pearl, Judea},
  year={2014},
  publisher={Elsevier}
}

@article{cen2025worldvla,
  title={Worldvla: Towards autoregressive action world model},
  author={Cen, Jun and Yu, Chaohui and Yuan, Hangjie and Jiang, Yuming and Huang, Siteng and Guo, Jiayan and Li, Xin and Song, Yibing and Luo, Hao and Wang, Fan and others},
  journal={arXiv preprint arXiv:2506.21539},
  year={2025}
}

@misc{black2024pi0,
      title={$\pi_0$: A Vision-Language-Action Flow Model for General Robot Control},
      author={Kevin Black and Noah Brown and Danny Driess and Adnan Esmail and Michael Equi and Chelsea Finn and Niccolo Fusai and Lachy Groom and Karol Hausman and Brian Ichter and Szymon Jakubczak and Tim Jones and Liyiming Ke and Sergey Levine and Adrian Li-Bell and Mohith Mothukuri and Suraj Nair and Karl Pertsch and Lucy Xiaoyang Shi and James Tanner and Quan Vuong and Anna Walling and Haohuan Wang and Ury Zhilinsky},
      year={2024},
      eprint={2410.24164},
      archivePrefix={arXiv},
      primaryClass={cs.LG},
      url={https://arxiv.org/abs/2410.24164},
}

@misc{black2025pi05,
      title={$\pi_{0.5}$: a Vision-Language-Action Model with Open-World Generalization},
      author={Physical Intelligence and Kevin Black and Noah Brown and James Darpinian and Karan Dhabalia and Danny Driess and Adnan Esmail and Michael Equi and Chelsea Finn and Niccolo Fusai and Manuel Y. Galliker and Dibya Ghosh and Lachy Groom and Karol Hausman and Brian Ichter and Szymon Jakubczak and Tim Jones and Liyiming Ke and Devin LeBlanc and Sergey Levine and Adrian Li-Bell and Mohith Mothukuri and Suraj Nair and Karl Pertsch and Allen Z. Ren and Lucy Xiaoyang Shi and Laura Smith and Jost Tobias Springenberg and Kyle Stachowicz and James Tanner and Quan Vuong and Homer Walke and Anna Walling and Haohuan Wang and Lili Yu and Ury Zhilinsky},
      year={2025},
      eprint={2504.16054},
      archivePrefix={arXiv},
      primaryClass={cs.LG},
      url={https://arxiv.org/abs/2504.16054},
}

@InProceedings{reuss2025flower,
  title = 	 {FLOWER: Democratizing Generalist Robot Policies with Efficient Vision-Language-Flow Models},
  author =       {Reuss, Moritz and Zhou, Hongyi and R\"{u}hle, Marcel and Ya\u{g}murlu, \"{O}mer Erdin\c{c} and Otto, Fabian and Lioutikov, Rudolf},
  booktitle = 	 {Proceedings of The 9th Conference on Robot Learning},
  pages = 	 {3736--3761},
  year = 	 {2025},
  editor = 	 {Lim, Joseph and Song, Shuran and Park, Hae-Won},
  volume = 	 {305},
  series = 	 {Proceedings of Machine Learning Research},
  month = 	 {27--30 Sep},
  publisher =    {PMLR},
  url = 	 {https://proceedings.mlr.press/v305/reuss25a.html},
}

@article{nvidia2024groot,
  title={Gr00t n1: An open foundation model for generalist humanoid robots},
  author={Bjorck, Johan and Casta{\~n}eda, Fernando and Cherniadev, Nikita and Da, Xingye and Ding, Runyu and Fan, Linxi and Fang, Yu and Fox, Dieter and Hu, Fengyuan and Huang, Spencer and others},
  journal={arXiv preprint arXiv:2503.14734},
  year={2025}
}

@article{shukor2025smolvla,
  title={Smolvla: A vision-language-action model for affordable and efficient robotics},
  author={Shukor, Mustafa and Aubakirova, Dana and Capuano, Francesco and Kooijmans, Pepijn and Palma, Steven and Zouitine, Adil and Aractingi, Michel and Pascal, Caroline and Russi, Martino and Marafioti, Andres and others},
  journal={arXiv preprint arXiv:2506.01844},
  year={2025}
}

@article{li2024cogact,
  title={Cogact: A foundational vision-language-action model for synergizing cognition and action in robotic manipulation},
  author={Li, Qixiu and Liang, Yaobo and Wang, Zeyu and Luo, Lin and Chen, Xi and Liao, Mozheng and Wei, Fangyun and Deng, Yu and Xu, Sicheng and Zhang, Yizhong and others},
  journal={arXiv preprint arXiv:2411.19650},
  year={2024}
}

@article{yue2024deer,
  title={Deer-vla: Dynamic inference of multimodal large language models for efficient robot execution},
  author={Yue, Yang and Wang, Yulin and Kang, Bingyi and Han, Yizeng and Wang, Shenzhi and Song, Shiji and Feng, Jiashi and Huang, Gao},
  journal={Advances in Neural Information Processing Systems},
  volume={37},
  pages={56619--56643},
  year={2024}
}

@inproceedings{zitkovich2023rt,
  title={Rt-2: Vision-language-action models transfer web knowledge to robotic control},
  author={Zitkovich, Brianna and Yu, Tianhe and Xu, Sichun and Xu, Peng and Xiao, Ted and Xia, Fei and Wu, Jialin and Wohlhart, Paul and Welker, Stefan and Wahid, Ayzaan and others},
  booktitle={Conference on Robot Learning},
  pages={2165--2183},
  year={2023},
  organization={PMLR}
}

@inproceedings{kim2025openvla,
  title={OpenVLA: An Open-Source Vision-Language-Action Model},
  author={Kim, Moo Jin and Pertsch, Karl and Karamcheti, Siddharth and Xiao, Ted and Balakrishna, Ashwin and Nair, Suraj and Rafailov, Rafael and Foster, Ethan P and Sanketi, Pannag R and Vuong, Quan and others},
  booktitle={Conference on Robot Learning},
  pages={2679--2713},
  year={2025},
  organization={PMLR}
}

@article{chi2023diffusion,
  title={Diffusion policy: Visuomotor policy learning via action diffusion},
  author={Chi, Cheng and Xu, Zhenjia and Feng, Siyuan and Cousineau, Eric and Du, Yilun and Burchfiel, Benjamin and Tedrake, Russ and Song, Shuran},
  journal={The International Journal of Robotics Research},
  volume={44},
  number={10-11},
  pages={1684--1704},
  year={2025},
  publisher={Sage Publications Sage UK: London, England}
}

@article{mees2022calvin,
  title={Calvin: A benchmark for language-conditioned policy learning for long-horizon robot manipulation tasks},
  author={Mees, Oier and Hermann, Lukas and Rosete-Beas, Erick and Burgard, Wolfram},
  journal={IEEE Robotics and Automation Letters},
  volume={7},
  number={3},
  pages={7327--7334},
  year={2022},
  publisher={IEEE}
}

@article{huang2025otter,
  title={Otter: A vision-language-action model with text-aware visual feature extraction},
  author={Huang, Huang and Liu, Fangchen and Fu, Letian and Wu, Tingfan and Mukadam, Mustafa and Malik, Jitendra and Goldberg, Ken and Abbeel, Pieter},
  journal={arXiv preprint arXiv:2503.03734},
  year={2025}
}

@article{dasari2024ditpi,
  title={The Ingredients for Robotic Diffusion Transformers},
  author = {Sudeep Dasari and Oier Mees and Sebastian Zhao and Mohan Kumar Srirama and Sergey Levine},
  journal = {arXiv preprint arXiv:2410.10088},
  year={2024},
}

@article{jiang2025lightvla,
  title={The better you learn, the smarter you prune: Towards efficient vision-language-action models via differentiable token pruning},
  author={Jiang, Titong and Jiang, Xuefeng and Ma, Yuan and Wen, Xin and Li, Bailin and Zhan, Kun and Jia, Peng and Liu, Yahui and Sun, Sheng and Lang, Xianpeng},
  journal={arXiv preprint arXiv:2509.12594},
  year={2025}
}

@article{ryoo2021tokenlearner,
  title={Tokenlearner: What can 8 learned tokens do for images and videos?},
  author={Ryoo, Michael S and Piergiovanni, AJ and Arnab, Anurag and Dehghani, Mostafa and Angelova, Anelia},
  journal={arXiv preprint arXiv:2106.11297},
  year={2021}
}

@article{rao2021dynamicvit,
  title={Dynamicvit: Efficient vision transformers with dynamic token sparsification},
  author={Rao, Yongming and Zhao, Wenliang and Liu, Benlin and Lu, Jiwen and Zhou, Jie and Hsieh, Cho-Jui},
  journal={Advances in neural information processing systems},
  volume={34},
  pages={13937--13949},
  year={2021}
}

@article{hou2025dita,
 title={Dita: Scaling Diffusion Transformer for Generalist Vision-Language-Action Policy},
 author={Hou, Zhi and Zhang, Tianyi and Xiong, Yuwen and Duan, Haonan and Pu, Hengjun and Tong, Ronglei and Zhao, Chengyang and Zhu, Xizhou and Qiao, Yu and Dai, Jifeng and Chen, Yuntao},
 journal={arXiv preprint arXiv:2503.19757},
 year={2025}
}

@article{reuss2024mdt,
  title={Multimodal diffusion transformer: Learning versatile behavior from multimodal goals},
  author={Reuss, Moritz and Ya{\u{g}}murlu, {\"O}mer Erdin{\c{c}} and Wenzel, Fabian and Lioutikov, Rudolf},
  journal={arXiv preprint arXiv:2407.05996},
  year={2024}
}

@misc{lipman2023flowmatching,
      title={Flow Matching for Generative Modeling},
      author={Yaron Lipman and Ricky T. Q. Chen and Heli Ben-Hamu and Maximilian Nickel and Matt Le},
      year={2023},
      eprint={2210.02747},
      archivePrefix={arXiv},
      primaryClass={cs.LG},
      url={https://arxiv.org/abs/2210.02747},
}

@article{liu2023libero,
  title={Libero: Benchmarking knowledge transfer for lifelong robot learning},
  author={Liu, Bo and Zhu, Yifeng and Gao, Chongkai and Feng, Yihao and Liu, Qiang and Zhu, Yuke and Stone, Peter},
  journal={Advances in Neural Information Processing Systems},
  volume={36},
  pages={44776--44791},
  year={2023}
}

@inproceedings{wu2024gr1,
  title={Unleashing Large-Scale Video Generative Pre-training for Visual Robot Manipulation},
  author={Hongtao Wu and Ya Jing and Chilam Cheang and Guangzeng Chen and Jiafeng Xu and Xinghang Li and Minghuan Liu and Hang Li and Tao Kong},
  booktitle={The Twelfth International Conference on Learning Representations},
  year={2024},
  url={https://openreview.net/forum?id=NxoFmGgWC9}
}

@article{li2023roboflamingo,
  title     = {Vision-Language Foundation Models as Effective Robot Imitators},
  author    = {Li, Xinghang and Liu, Minghuan and Zhang, Hanbo and Yu, Cunjun and Xu, Jie and Wu, Hongtao and Cheang, Chilam and Jing, Ya and Zhang, Weinan and Liu, Huaping and Li, Hang and Kong, Tao},
  journal={arXiv preprint arXiv:2311.01378},
  year={2023}
}

@inproceedings{black2024susie,
  title={Zero-Shot Robotic Manipulation with Pre-Trained Image-Editing Diffusion Models},
  author={Kevin Black and Mitsuhiko Nakamoto and Pranav Atreya and Homer Rich Walke and Chelsea Finn and Aviral Kumar and Sergey Levine},
  booktitle={The Twelfth International Conference on Learning Representations},
  year={2024},
  url={https://openreview.net/forum?id=c0chJTSbci}
}

@inproceedings{hu2024vpp,
  title={Video Prediction Policy: A Generalist Robot Policy with Predictive Visual Representations},
  author={Hu, Yucheng and Guo, Yanjiang and Wang, Pengchao and Chen, Xiaoyu and Wang, Yen-Jen and Zhang, Jianke and Sreenath, Koushil and Lu, Chaochao and Chen, Jianyu},
  booktitle={Forty-second International Conference on Machine Learning},
  year={2025}
}

@article{tian2024predictive,
  title={Predictive inverse dynamics models are scalable learners for robotic manipulation},
  author={Tian, Yang and Yang, Sizhe and Zeng, Jia and Wang, Ping and Lin, Dahua and Dong, Hao and Pang, Jiangmiao},
  journal={arXiv preprint arXiv:2412.15109},
  year={2024}
}

@misc{kim2025oft,
      title={Fine-Tuning Vision-Language-Action Models: Optimizing Speed and Success},
      author={Moo Jin Kim and Chelsea Finn and Percy Liang},
      year={2025},
      eprint={2502.19645},
      archivePrefix={arXiv},
      primaryClass={cs.RO},
      url={https://arxiv.org/abs/2502.19645},
}

@misc{li2025coavla,
  title={CoA-VLA: Improving Vision-Language-Action Models via Visual-Textual Chain-of-Affordance},
  author={Jinming Li and Yichen Zhu and Zhibin Tang and Junjie Wen and Minjie Zhu and Xiaoyu Liu and Chengmeng Li and Ran Cheng and Yaxin Peng and Yan Peng and Feifei Feng},
  year={2025},
  eprint={2412.20451},
  archivePrefix={arXiv},
  primaryClass={cs.RO},
  url={https://arxiv.org/abs/2412.20451},
}

@misc{wu2023gello,
  title={GELLO: A General, Low-Cost, and Intuitive Teleoperation Framework for Robot Manipulators},
  author={Philipp Wu and Yide Shentu and Zhongke Yi and Xingyu Lin and Pieter Abbeel},
  year={2024},
  eprint={2309.13037},
  archivePrefix={arXiv},
  primaryClass={cs.RO},
  url={https://arxiv.org/abs/2309.13037},
}

\appendix

\providecommand{\na}{--}

\section{Noisy-OR Relaxation for Mask Union}
\label{app:selector-preliminaries}

This appendix expands the noisy-OR construction inlined in \cref{sec:method-integration}: how it arises as the expected hard mask under a natural row-wise voting model, why it is preferred to simpler differentiable surrogates, and how it interacts with the diversity regularizer. The instantiations used by the selector — across voting rows within a branch and across branches — are stated in the main text and are not repeated here.

\paragraph{Boolean union and its noisy-OR relaxation.}
The Boolean union of events $b_1,\ldots,b_n\in\{0,1\}$,
\begin{equation}
  b_\vee = \bigvee_{i=1}^{n} b_i = \mathbf{1}\!\left[\textstyle\sum_{i=1}^{n} b_i > 0\right],
\end{equation}
is non-differentiable. Replacing each $b_i$ by a Bernoulli probability $p_i\in[0,1]$ and assuming the events are conditionally independent gives the differentiable surrogate
\begin{equation}
  \rho_\vee(p_1,\ldots,p_n) = \Pr\!\left(\bigvee_{i=1}^{n} b_i = 1\right) = 1-\prod_{i=1}^{n}(1-p_i),
\end{equation}
known as the \emph{noisy-OR}~\cite{pearl1988probabilistic}. The term $\prod_i(1-p_i)$ is the probability that no event occurs, so subtracting it from one gives the probability that at least one does.

\paragraph{Probabilistic interpretation in the selector.}
The noisy-OR is not introduced from the outside as a smooth approximation of $\vee$. Treating each row's softmax $P_\omega[i,\cdot]$ as a categorical distribution over which token row $i$ votes for, and the rows as conditionally independent, the probability that token $j$ receives at least one vote across the $N_v$ rows is exactly
\begin{equation}
  \mathbb{E}\!\left[m_\omega[j]\right]
  \;=\;
  \Pr\!\left(\exists\, i:\, z_\omega[i]=j\right)
  \;=\;
  1 - \prod_{i=1}^{N_v}\!\left(1-P_\omega[i,j]\right)
  \;=\;
  \rho_\omega[j],
\end{equation}
where $m_\omega[j] = \mathbf{1}[\exists\, i:\, z_\omega[i]=j]$ is the hard branch mask. The same identity applies across branches when branch decisions are taken as independent. The soft scores $\rho_\omega[j]$ and $\rho[j]$ used in \cref{sec:method-integration} are therefore the expected hard masks under the row-wise and cross-branch surrogate voting models, not arbitrary smooth surrogates.

\paragraph{Properties relied on by the selector.}
\begin{itemize}[leftmargin=*,itemsep=2pt,topsep=2pt]
  \item \textbf{(P1) Boundary agreement with hard OR.} $\rho_\vee = 0$ iff every $p_i = 0$, and $\rho_\vee \to 1$ if any $p_i \to 1$. The forward-pass mask and the soft expectation agree at the corners of $[0,1]^n$.
  \item \textbf{(P2) Strict monotonicity with non-degenerate gradient.} $\partial \rho_\vee / \partial p_i = \prod_{k\ne i}(1-p_k) \geq 0$, with equality only when some other $p_k = 1$. Every input receives a gradient unless the union is already saturated by another event.
  \item \textbf{(P3) Bounded without renormalization.} $\rho_\vee \in [0,1]$ for any $n$, so evidence aggregated over $N_v$ rows or two branches needs no division by $n$.
  \item \textbf{(P4) Union-style accumulation without dilution or overshoot.} Two independent moderate votes reinforce (e.g.\ $p_1=p_2=0.5$ give $\rho_\vee = 0.75$), a single confident vote keeps the union near $1$ regardless of how many low-confidence votes accompany it, and the value never exceeds $1$.
\end{itemize}

\paragraph{Comparison with simpler surrogates.}
\Cref{tab:noisy-or-alternatives} contrasts noisy-OR with three natural alternatives. The maximum $\max_i p_i$ satisfies P1 and P3 but is gradient-zero away from the $\arg\max$ (P2 partial) and does not accumulate: two votes at $p=0.5$ stay at $0.5$ rather than reinforcing each other (P4 violated). The sum $\sum_i p_i$ violates P1 and P3 by exceeding $1$. The normalized mean $\tfrac{1}{n}\sum_i p_i$ restores boundedness but breaks P1, and \emph{dilutes} one confident vote among many low-confidence ones down toward $1/n$ — exactly the regime the selector operates in, where only a small fraction of rows vote confidently for any given column.

\begin{table}[t]
  \centering
  \caption{Differentiable surrogates for the Boolean union of $n$ Bernoulli events. Only the noisy-OR satisfies all four properties.}
  \label{tab:noisy-or-alternatives}
  \small
  \setlength{\tabcolsep}{8pt}
  \begin{tabular}{@{}lcccc@{}}
    \toprule
    \textbf{Surrogate} & \textbf{P1} & \textbf{P2} & \textbf{P3} & \textbf{P4} \\
    \midrule
    $\max_i p_i$                          & Yes & Partial & Yes & No  \\
    $\sum_i p_i$                          & No  & Yes     & No  & No  \\
    $\tfrac{1}{n}\sum_i p_i$              & No  & Yes     & Yes & No  \\
    \textbf{Noisy-OR} $1-\prod_i(1-p_i)$  & Yes & Yes     & Yes & Yes \\
    \bottomrule
  \end{tabular}
\end{table}

\paragraph{Independence assumption and the role of the diversity regularizer.}
The expected-mask identity above holds exactly when the underlying votes are conditionally independent; otherwise the noisy-OR is an approximation. In the selector this assumption appears at two scales.

\emph{Across voting rows within a branch.} The rows share a single score matrix and are not strictly independent, but each row casts at most one effective vote and the guidance-conditioned queries separate row distributions, so we expect the residual dependence to be limited.

\emph{Across branches.} We do not attempt to enforce probabilistic independence between the language and proprioception branches. The diversity regularizer $\mathcal{L}_{\mathrm{div}}$ instead discourages branch collapse by penalizing cosine similarity between the row-aggregated selection distributions $\softmax(r_{\mathrm{lang}})$ and $\softmax(r_{\mathrm{prop}})$, which reduces empirical overlap and keeps branch behavior aligned with the complementary-branch intent of the architecture. This narrows the regime in which the cross-branch noisy-OR is loosest — heavily overlapping branches — without claiming the approximation is exact.

\paragraph{Role in the forward pass.}
The noisy-OR never enters the forward computation. The selector uses the hard Boolean union $m[j] = m_{\mathrm{lang}}[j] \vee m_{\mathrm{prop}}[j]$ both at inference and at the forward step of training, and the straight-through gate routes gradients through the soft $\rho[j]$ while leaving the forward value equal to $m[j]$. The relaxation therefore only shapes the learning signal; deployed model behavior is identical to that of a deterministic argmax-union selector.

\paragraph{Gradient flow and token exploration.}
The straight-through gate passes gradient only to tokens in the current selection $M$: tokens with $m[j]=0$ are dropped from $H_v^M$ before the VLM, so their gate value $w[j]$ never enters the loss and they receive no direct gradient in that step. Coverage of the full token set is instead provided over training by the Gumbel perturbation $\hat S_\omega = S_\omega + \alpha\Gamma_\omega$. Early in training, the large exploration scale $\alpha$ makes the per-row argmax votes stochastic, so the realized selection $M$ varies from step to step and, across steps, most visual tokens are eventually selected and updated at least once. A dropped token also continues to shape learning indirectly, since it participates in the scores of other tokens (as a voting row in $\tilde H_v$ and through the row-wise normalization) and in the aggregated selection distributions $r_\omega$ used by $\mathcal{L}_{\mathrm{div}}$. As $\alpha$ anneals from $1.0$ to $0.01$, selection sharpens toward the deterministic argmax-union used at inference, so exploration is concentrated early and the gate becomes effectively hard by convergence. This is distinct from property~(P2), which concerns the gradient of the noisy-OR surrogate with respect to its probability inputs rather than the realization of the discrete forward selection.

\section{Experiment Details}
\label{app:exp-details}

\subsection{Observations and Proprioceptive State}
\label{app:observations-proprio}
Image preprocessing follows the dataset transform configs. For CALVIN, the static and gripper-camera views are resized to $224\times224$. During training, we apply RandomShiftsAug with padding 10 for the static view and padding 4 for the gripper view, then scale images to $[0,1]$ and normalize with CLIP statistics. Validation disables RandomShiftsAug but keeps resize and normalization. The proprioceptive state is the 15D robot observation vector, normalized using dataset statistics and with additional normalization of orientation entries. Actions are 7D relative end-effector commands scaled to $[-1,1]$.

For LIBERO, we use the same augmentation and normalization structure, resizing images to $112\times112$. We form a 9D proprioceptive state by concatenating seven joint values with the 2D gripper state. Actions use the same 7D relative command parameterization and are scaled to $[-1,1]$.

For real-world evaluation, we use the same two-stream policy interface with a fixed third-person camera and a wrist-mounted camera. Both views are resized to $224\times224$ and normalized with Florence/CLIP-style image statistics; training additionally uses RandomShiftsAug with padding 10 for the static view, padding 4 for the wrist view, and color jitter. The UR3 proprioceptive state is a 16D vector containing end-effector position (3), end-effector quaternion (4), gripper state (1), six joint values (6), and two zero-padded dimensions, normalized using statistics computed from training episodes. Actions contain six relative pose deltas and one gripper command; the motion dimensions are normalized and the gripper target is binarized. Across CALVIN, LIBERO, and real-world runs, unless otherwise specified, normalized proprioceptive values are clipped to $[-3,3]$, uniformly discretized into $B=256$ bins, and mapped to VLM-vocabulary token IDs as in \cref{sec:method-encoding}; \cref{tab:bin-clip-sensitivity} studies this bin and clip choice. The real-world dataset contains approximately 120 minutes of teleoperated demonstrations over 14 tasks: 12 pick-and-place tasks across banana, corn, and tape using Basket$\rightarrow$Plate, Basket$\rightarrow$Table, Plate$\rightarrow$Basket, and Table$\rightarrow$Basket routes, plus two drawer tasks, open and close.

\begin{figure}[t]
  \centering
  \begin{subfigure}[t]{0.48\linewidth}
    \centering
    \includegraphics[width=\linewidth]{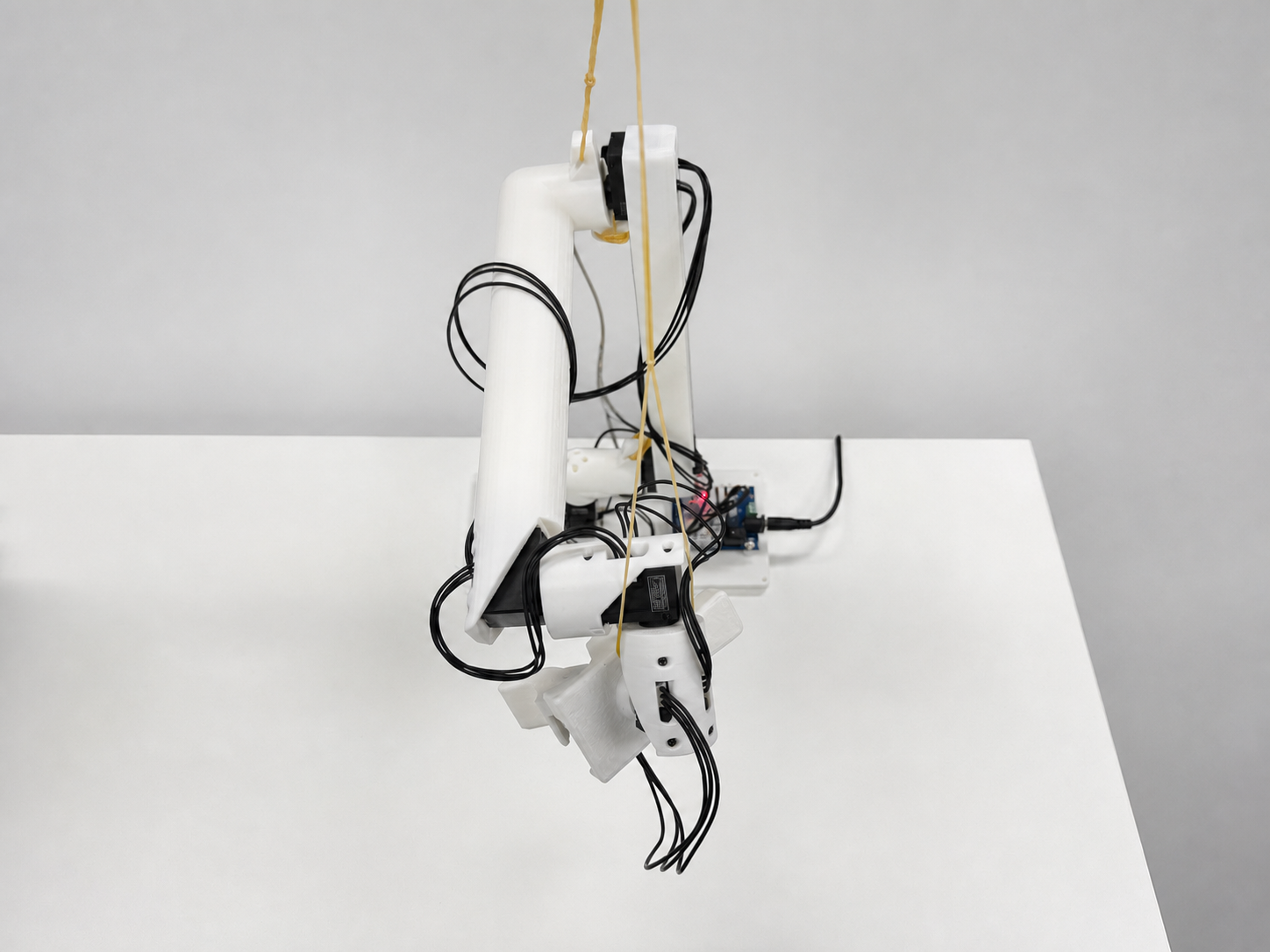}
    \caption{Overhead view.}
  \end{subfigure}
  \hfill
  \begin{subfigure}[t]{0.48\linewidth}
    \centering
    \includegraphics[width=\linewidth]{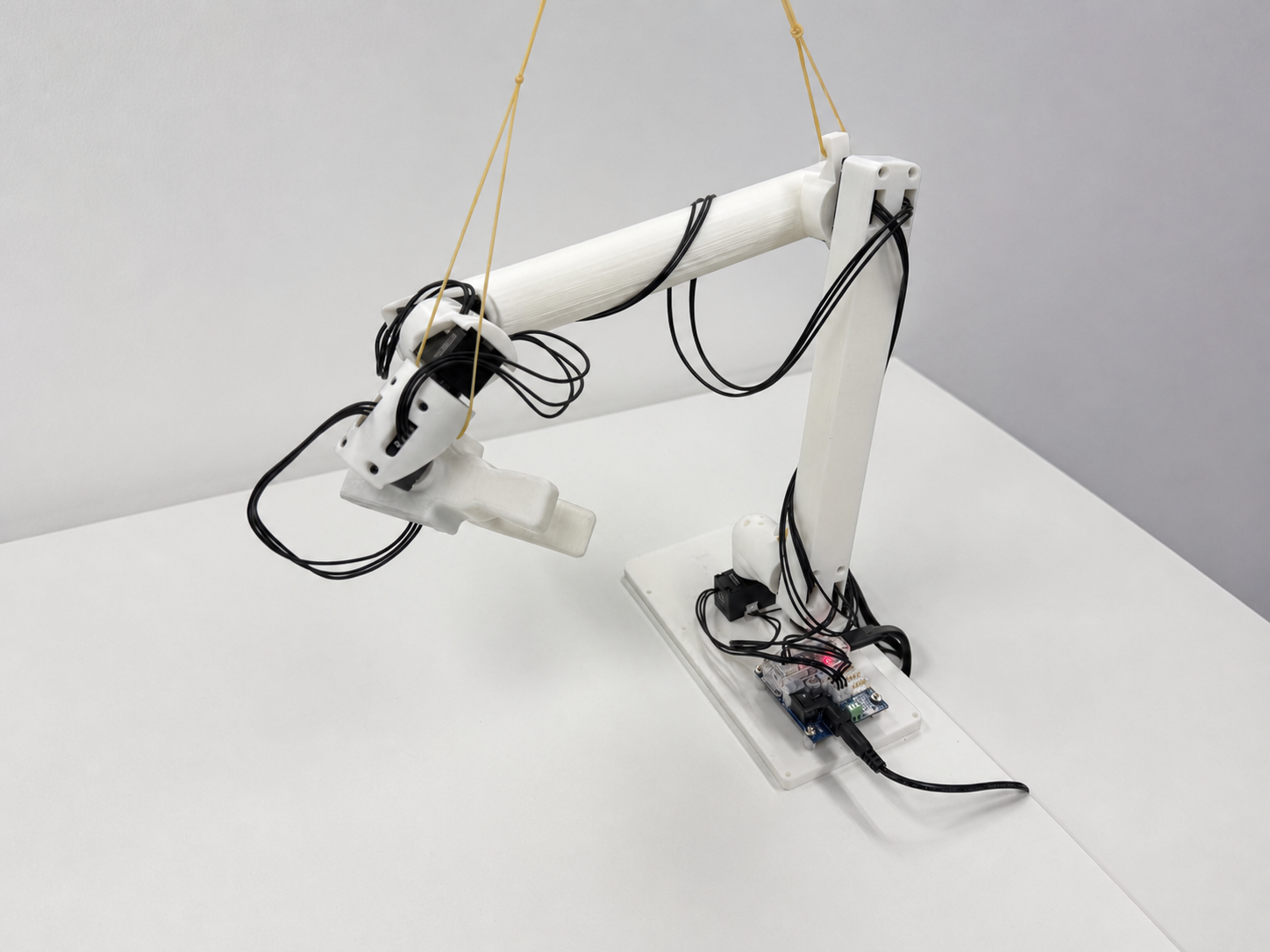}
    \caption{Side view.}
  \end{subfigure}
  \caption{Customized GELLO-style leader-side teleoperation interface used for collecting real-world demonstrations.}
  \label{fig:teleop-system}
\end{figure}

\paragraph{Teleoperation interface.}
The real-world demonstrations were collected with the customized GELLO-style leader-side teleoperation interface shown in \cref{fig:teleop-system}~\citep{wu2023gello}. The operator moves a lightweight linkage and gripper handle, and the adapted software maps these inputs to the UR3 during data collection. The policy observations remain the fixed and wrist-mounted RGB streams together with the robot proprioceptive state described above.

\subsection{Model Architecture}
\label{app:model-architecture}
We build on \textsc{FLOWER} and use Florence-2-Large (\texttt{microsoft/Florence-2-large}) as the vision-language backbone. Following FLOWER's recipe, we fine-tune Florence rather than freezing it, so comparisons use the same backbone adaptation regime. A special token \texttt{<Flow>} is embedded and inserted to mark the conditioning boundary, and we apply token dropout with probability 0.1 to the VLM encoder outputs during training. The pre-VLM selector uses the branch-specific vote-matrix scoring described in \cref{sec:method-integration} for the instruction and proprioception branches, and its global context token uses the learned linear projection $W_c$ to compute $H^{\mathrm{ctx}}$ from the mean pre-selection visual feature. The action generator is a rectified-flow / Diffusion Transformer with hidden size 1024, 18 transformer layers, and 16 attention heads, using dropout 0.1 in attention, residual, and MLP blocks. The policy predicts an action chunk of length 10, executed with chunked replanning every 10 environment steps.

\subsection{Optimization Hyperparameters}
We train end-to-end with AdamW using learning rate $2\times 10^{-5}$, betas $(0.9,0.95)$, and weight decay 0.05 applied to non-normalization and non-bias parameters (norm/bias parameters use zero weight decay). Training uses bf16-mixed precision with per-GPU batch size 8 on one GPU. We use a tri-stage learning-rate schedule over 50k steps (matching \texttt{max\_epochs=50} and \texttt{limit\_train\_batches=1000}): (i) linear warmup for 5\% of steps from $0.1\times\mathrm{lr}$ to $\mathrm{lr}$, (ii) constant hold for 10\% of steps, and (iii) cosine decay for the remaining 85\% to a final learning rate of $0.5\times\mathrm{lr}$. Selector training anneals the Gumbel exploration scale from 1.0 to 0.01 and uses the branch diversity coefficient $\lambda_{\mathrm{div}}=5\times10^{-4}$ from the objective $\mathcal{L}=\mathcal{L}_{\mathrm{fm}}+\lambda_{\mathrm{div}}\mathcal{L}_{\mathrm{div}}$, with margin $\gamma=0.75$. We additionally maintain an exponential moving average (EMA) of parameters with decay 0.999 and use EMA weights for evaluation.

\begin{table}[t]
  \centering
  \setlength{\tabcolsep}{6pt}
  \caption{Training and optimization hyperparameters used across experiments.}
  \begin{tabular}{l l}
    \toprule
    Setting & Value \\
    \midrule
    Optimizer & AdamW \\
    Learning rate & $2\times 10^{-5}$ \\
    AdamW betas & $(0.9,0.95)$ \\
    Weight decay & $0.05$ (no decay on norm/bias) \\
    Per-GPU batch size & $8$ \\
    Number of GPUs & $1$ \\
    Total steps & $50{,}000$ \\
    LR schedule & tri-stage (warmup $\rightarrow$ hold $\rightarrow$ cosine) \\
    Warmup / hold / decay & $0.05 / 0.10 / 0.85$ \\
    Warmup start lr & $0.1\times \mathrm{lr}$ \\
    Final lr & $0.5\times \mathrm{lr}$ \\
    Precision & bf16-mixed \\
    EMA decay & $0.999$ \\
    VLM token dropout & $0.1$ \\
    Gumbel scale & $1.0 \rightarrow 0.01$ \\
    Diversity loss & $\lambda_{\mathrm{div}}=5\times10^{-4}$, margin $\gamma=0.75$ \\
    \bottomrule
  \end{tabular}
\end{table}

\subsection{Inference Settings and Measurement Protocol}
\label{app:latency-vram}
At inference time, we run rectified-flow sampling with 4 steps (\texttt{num\_sampling\_steps=4}) per action chunk. The model predicts a 10-step chunk and replans every 10 environment steps. Inference is performed in bf16 and always uses both camera views.

CALVIN long-horizon evaluation follows the standard 5-subtask instruction-chain protocol: each subtask is given up to 360 environment steps, and we report success rates for completing 1 through 5 subtasks as well as the average successful sequence length. LIBERO evaluation uses 50 trials per task with a maximum horizon of 520 environment steps; we report per-task success and suite averages.

For latency and VRAM profiling, we measure per-environment-step end-to-end inference time including vision encoding, the pre-VLM selector, the Florence encoder forward pass, and all diffusion sampling steps. Although the policy predicts 10-action chunks, the latency values in \cref{tab:efficiency,tab:latency_breakdown} are reported per environment step rather than as a total per action chunk. We compute latency with CUDA synchronization to avoid asynchronous kernel overlap artifacts, and report the mean over a fixed number of inference steps (1000 steps in our efficiency tables). In \cref{tab:latency_breakdown}, the Vision column includes encoding both camera views. Peak VRAM is reported as the maximum allocated GPU memory during inference. All profiling numbers in the paper are collected on a single RTX~4090 GPU under the same bf16 and two-view settings as evaluation.

\subsection{Baseline Details}
\label{app:baseline-details}

For the extended tables, we keep only the architecture attributes needed to interpret efficiency: scale, backbone, and visual-token count; ``--'' denotes not applicable or not reported.

\noindent\textbf{FLOWER.} FLOWER \citep{reuss2025flower} is the closest dual-system baseline to our implementation. It uses Florence-2-Large as the vision-language backbone and conditions a flow-based action generator through dense cross-attention. We compare against FLOWER under the same benchmark settings used in the main simulation table.

\noindent\textbf{$\pi_0$ and $\pi_{0.5}$.} $\pi_0$ \citep{black2024pi0} and $\pi_{0.5}$ \citep{black2025pi05} pair a PaliGemma VLM with a flow-based action generator. We fine-tune the open variants on CALVIN ABC$\to$D, using them to compare continuous proprioceptive action-head conditioning against token-form proprioceptive input. These are the $^*$-marked $\pi_0$ and $\pi_{0.5}$ rows in the main CALVIN table.

\noindent\textbf{OpenVLA and OpenVLA-OFT.} OpenVLA \citep{kim2025openvla} is a single-system VLA that discretizes actions as language-model tokens. OpenVLA-OFT \citep{kim2025oft} extends this family with an action expert and improved fine-tuning recipe, and is included for LIBERO comparisons.

\noindent\textbf{GR-1.} GR-1 \citep{wu2024gr1} is a generative robot policy that combines visual encoders with a sequence model for language-conditioned manipulation. We include it as a compact VLA-style baseline on CALVIN.

\noindent\textbf{RoboFlamingo and DeerVLA.} RoboFlamingo \citep{li2023roboflamingo} adapts a Flamingo-style vision-language model to robot control. DeerVLA \citep{yue2024deer} builds on this family with improved data and training choices, providing additional single-system VLA comparisons on CALVIN.

\noindent\textbf{LightVLA.} LightVLA \citep{jiang2025lightvla} is the closest efficiency-oriented baseline. We distinguish it from \ours{} along four axes:
\begin{itemize}
  \item \textbf{Target.} LightVLA targets visual-token computation reduction through instruction-guided pruning, whereas \ours{} asks how proprioception should participate in visual grounding.
  \item \textbf{Guidance signal.} LightVLA uses the instruction as the pruning signal; \ours{} uses separate language and proprioception branches so task semantics and robot-state cues can make complementary votes.
  \item \textbf{Selection mechanism.} \ours{} merges branch masks with a hard union in the forward pass and a noisy-OR relaxation for gradients.
  \item \textbf{Regularization and context.} \ours{} uses a diversity loss to discourage the two branches from collapsing to the same selection pattern, and keeps a global context token so aggressive local selection does not discard coarse scene information.
\end{itemize}

\noindent\textbf{Diff-P-CNN and MDT.} Diff-P-CNN \citep{chi2023diffusion} and MDT \citep{reuss2024mdt} are diffusion-style manipulation policies without a large VLM backbone. They provide non-VLM references for CALVIN, especially on the D$\to$D split.

\noindent\textbf{SuSIE, VPP, and Seer.} SuSIE \citep{black2024susie}, VPP \citep{hu2024vpp}, and Seer \citep{tian2024predictive} are visual planning or predictive representation baselines. They are included to compare against methods that rely on planning-oriented visual abstractions rather than a VLM-action-head decomposition.

\noindent\textbf{RoboUniView.} RoboUniView is included in the CALVIN D$\to$D comparison as a prior multi-view robot learning baseline. We report it only where the corresponding split result is available.

\section{Additional Results}
\label{app:additional-results}

\subsection{CALVIN Long-Horizon Results}
\label{app:calvin-long-horizon-results}

\Cref{tab:calvin:task_ABC_D_extended} extends the main CALVIN ABC$\to$D comparison in \cref{tab:main-sim}a with architecture details, while \cref{tab:calvin:task_ABCD_D,tab:calvin:task_D_D} report the additional ABCD$\to$D and D$\to$D splits. In these tables, the Tokens column is the average number of retained visual tokens per timestep out of 100 input visual tokens. The split-specific means for \ours{} are 12, 15, and 14 tokens; the ``around \SI{15}{\percent}'' statement in the main text is a rounded summary across CALVIN settings.

\begin{table}[ht]
  \centering
  \caption{CALVIN ABC$\to$D long-horizon success with architecture details.}
  \label{tab:calvin:task_ABC_D_extended}
  \small
  \resizebox{\columnwidth}{!}{%
    \begin{tabular}{@{}l c l c
        S[table-format=2.1, detect-weight]
        S[table-format=2.1, detect-weight]
        S[table-format=2.1, detect-weight]
        S[table-format=2.1, detect-weight]
        S[table-format=2.1, detect-weight]
      S[table-format=1.2, detect-weight]@{}}
      \toprule
      \textbf{Method} &
      \multicolumn{3}{c}{\textbf{Architecture}} &
      \multicolumn{6}{c}{\textbf{Performance}} \\
      \cmidrule(lr){2-4}\cmidrule(lr){5-10}
      & \textbf{Scale (B)} & \textbf{Backbone} & {\textbf{Tokens} $\downarrow$} &
      {\textbf{LH-1} $\uparrow$} & {\textbf{LH-2} $\uparrow$} & {\textbf{LH-3} $\uparrow$} & {\textbf{LH-4} $\uparrow$} & {\textbf{LH-5} $\uparrow$} & {\textbf{Avg. Len.} $\uparrow$} \\
      \midrule
      SuSIE             & \na  & \na & \na & 87.0 & 69.0 & 49.0 & 38.0 & 26.0 & 2.69 \\
      VPP               & 1.5  & SVD+CLIP & \na & 95.7 & 91.2 & 86.3 & 81.0 & 75.0 & 4.29 \\
      Seer              & 0.3  & ViT+CLIP & \na & 96.3 & 91.6 & 86.1 & 80.3 & 74.0 & 4.29 \\
      OpenVLA           & 7.7  & \shortstack[l]{Llama-2\\DINOv2+SigLIP} & 256 & 91.3 & 77.8 & 62.0 & 52.1 & 43.5 & 3.27 \\
      GR-1              & 0.195  & \shortstack[l]{CLIP\\MAE-ViT} & \na & 85.4 & 71.2 & 59.6 & 49.7 & 40.1 & 3.06 \\
      RoboFlamingo      & 3  & \shortstack[l]{OpenFlamingo\\ViT} & 128 & 82.4 & 61.9 & 46.6 & 33.1 & 23.5 & 2.47 \\
      $\pi_0^{*}$       & 3.3  & PaliGemma & 512 & 70.0 & 48.0 & 37.0 & 28.0 & 18.0 & 2.01 \\
      $\pi_{0.5}^{*}$   & 3.3  & PaliGemma & 512 & 71.0 & 56.0 & 45.0 & 37.0 & 29.0 & 2.38 \\
      FLOWER            & 0.95 & Florence-2-L & 100 & \bfseries 99.3 & \bfseries 96.0 & \underline{90.3} & \underline{82.3} & \underline{75.5} & \underline{4.44} \\
      \midrule
      \textbf{\ours{}}  & 0.95 & Florence-2-L & \bfseries 12 & \underline{98.9} & \underline{95.4} & \bfseries 91.6 & \bfseries 86.4 & \bfseries 79.1 & \bfseries 4.52 \\
      \bottomrule
    \end{tabular}
  }
\end{table}

\begin{table}[ht]
  \centering
  \caption{CALVIN ABCD$\to$D long-horizon success with architecture details.}
  \label{tab:calvin:task_ABCD_D}
  \small
  \resizebox{\columnwidth}{!}{%
    \begin{tabular}{@{}l c l c
        S[table-format=2.2, detect-weight]
        S[table-format=2.1, detect-weight]
        S[table-format=2.1, detect-weight]
        S[table-format=2.1, detect-weight]
        S[table-format=2.1, detect-weight]
      S[table-format=1.2, detect-weight]@{}}
      \toprule
      \textbf{Method} &
      \multicolumn{3}{c}{\textbf{Architecture}} &
      \multicolumn{6}{c}{\textbf{Performance}} \\
      \cmidrule(lr){2-4}\cmidrule(lr){5-10}
      & \textbf{Scale (B)} & \textbf{Backbone} & {\textbf{Tokens} $\downarrow$} &
      {\textbf{LH-1} $\uparrow$} & {\textbf{LH-2} $\uparrow$} & {\textbf{LH-3} $\uparrow$} & {\textbf{LH-4} $\uparrow$} & {\textbf{LH-5} $\uparrow$} & {\textbf{Avg. Len.} $\uparrow$} \\
      \midrule
      Diff-P-CNN     & 0.32  & \na & \na & 86.3 & 72.7 & 60.1 & 51.2 & 41.7 & 3.16 \\
      RoboFlamingo   & 3.0  & \shortstack[l]{OpenFlamingo\\ViT} & 128 & 96.4 & 89.6 & 82.4 & 74.0 & 66.0 & 4.09 \\
      DeerVLA        & 3.0  & \shortstack[l]{OpenFlamingo} & 128 & 99.1 & 93.3 & 82.1 & 74.6 & 63.8 & 4.13 \\
      GR-1           & 0.195  & \shortstack[l]{MAE-ViT\\GPT} & \na & 94.9 & 89.6 & 84.4 & 78.9 & 73.1 & 4.21 \\
      FLOWER         & 0.95 & Florence-2-L & 100 & 98.9 & 96.7 & 93.9 & 90.2 & 85.5 & 4.62 \\
      \midrule
      \textbf{\ours{}} & 0.95 & Florence-2-L & \bfseries 15 &
      \multicolumn{1}{c}{\bfseries 99.5} & \multicolumn{1}{c}{\bfseries 97.2} & \multicolumn{1}{c}{\bfseries 96.6} & \multicolumn{1}{c}{\bfseries 92.3} & \multicolumn{1}{c}{\bfseries 88.5} & \multicolumn{1}{c}{\bfseries 4.74} \\
      \bottomrule
    \end{tabular}
  }
\end{table}

\begin{table}[ht]
  \centering
  \caption{CALVIN D$\to$D long-horizon success with architecture details.}
  \label{tab:calvin:task_D_D}
  \small
  \resizebox{\columnwidth}{!}{%
    \begin{tabular}{@{}l c l c
        S[table-format=2.1, detect-weight]
        S[table-format=2.1, detect-weight]
        S[table-format=2.1, detect-weight]
        S[table-format=2.1, detect-weight]
        S[table-format=2.1, detect-weight]
      S[table-format=1.2, detect-weight]@{}}
      \toprule
      \textbf{Method} &
      \multicolumn{3}{c}{\textbf{Architecture}} &
      \multicolumn{6}{c}{\textbf{Performance}} \\
      \cmidrule(lr){2-4}\cmidrule(lr){5-10}
      & \textbf{Scale (B)} & \textbf{Backbone} & {\textbf{Tokens} $\downarrow$} &
      {\textbf{LH-1} $\uparrow$} & {\textbf{LH-2} $\uparrow$} & {\textbf{LH-3} $\uparrow$} & {\textbf{LH-4} $\uparrow$} & {\textbf{LH-5} $\uparrow$} & {\textbf{Avg. Len.} $\uparrow$} \\
      \midrule
      MDT & \na & \na & \na & 93.7 & 84.5 & 74.1 & 64.4 & 55.6 & 3.72 \\
      RoboUniView & \na & \na & \na & 96.2 & 88.8 & 77.6 & 66.6 & 56.3 & 3.85 \\
      \textbf{\ours{}} & 0.95 & Florence-2-L & \bfseries 14 & \bfseries 96.9 & \bfseries 89.8 & \bfseries 83.6 & \bfseries 80.5 & \bfseries 72.7 & \bfseries 4.23 \\
      \bottomrule
    \end{tabular}
  }
\end{table}

LH-$k$ is the success rate (\%) of completing $k$ consecutive subtasks in the five-subtask evaluation chain, and Avg.\ Len.\ is the mean number of consecutively completed subtasks. The architecture columns provide model scale, backbone, and visual-token count when available. In the ABCD$\to$D setting, \ours{} achieves the best Avg.\ Len.\ (4.74) and the best LH-5 result (88.5). In D$\to$D, \ours{} exceeds the remaining prior baselines reported in \Cref{tab:calvin:task_D_D}.

\subsection{CALVIN Subtask Breakdown}
\label{app:calvin-subtask-breakdown}

\begin{figure}[ht]
  \centering
  \includegraphics[width=\columnwidth,height=0.6\textheight,keepaspectratio]{fig/calvin\_task\_breakdown.pdf}
  \caption{Subtask-level breakdown of \ours{} on CALVIN ABC$\to$D. We plot the 12 lowest-success subtasks among the 34 CALVIN subtasks, ordered by failure rate. Bars report failure rate; right annotations report success rate with successful/evaluated rollout counts.}
  \label{fig:calvin_34subtasks}
\end{figure}

\Cref{fig:calvin_34subtasks} complements the chain-level CALVIN metrics by showing where the remaining ABC$\to$D errors occur. Most subtasks are near saturation, so residual failures concentrate in a small set of contact-sensitive or placement-sensitive interactions. Only three subtasks fall below $80\%$ success in this evaluation: \textit{Lift Pink Block Drawer} ($78.6\%$), \textit{Push Pink Block Right} ($78.8\%$), and \textit{Push Into Drawer} ($79.4\%$). The next hardest cases, such as \textit{Place In Slider} and \textit{Stack Block}, also require precise placement or sustained contact. This pattern is consistent with the main CALVIN result: \ours{} improves long-horizon chain completion, while the remaining failures are concentrated in interactions where small state-estimation or contact-control errors can accumulate. \Cref{app:qualitative-failure-cases} revisits the same contact-and-approach failure mode in real-world rollouts.

\subsection{Long-Horizon Qualitative Rollouts}
\label{app:long-horizon-qualitative-rollouts}

\begin{figure}[!htbp]
  \centering
  \includegraphics[width=\linewidth]{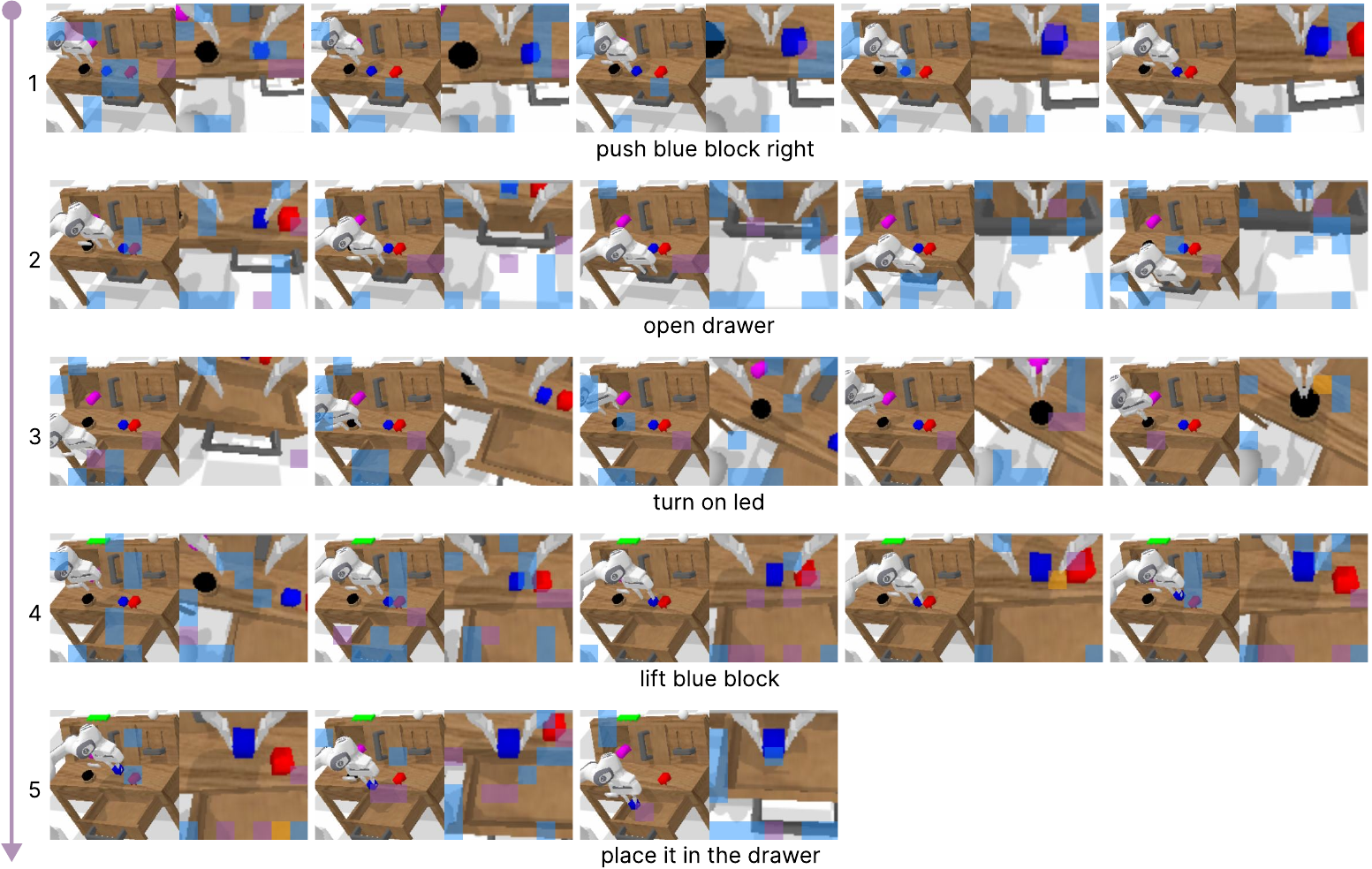}
  \caption{Qualitative CALVIN LH-5 rollout from \ours{}. Rows correspond to the five consecutive subtasks in the evaluation chain, and columns show representative timesteps within each subtask. Colored overlays visualize visual tokens retained by the instruction and proprioception guidance branches.}
  \label{fig:calvin-lh5-rollout}
\end{figure}

\begin{figure}[!htbp]
  \centering
  \includegraphics[width=\linewidth]{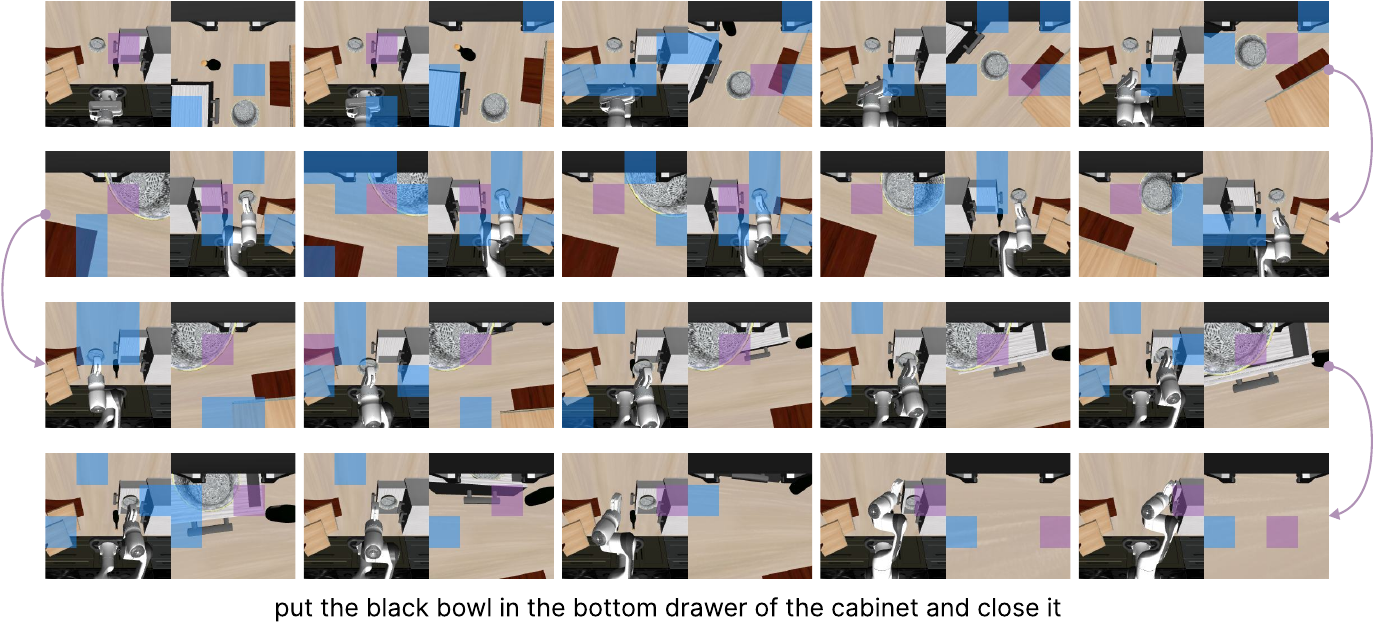}
  \caption{Qualitative LIBERO-Long rollout from \ours{} for placing the black bowl in the bottom drawer and closing it. Colored overlays visualize retained visual tokens as the policy transitions from reaching and grasping to drawer interaction and closing.}
  \label{fig:libero-long-rollout}
\end{figure}

\clearpage

\Cref{fig:calvin-lh5-rollout,fig:libero-long-rollout} provide qualitative examples of the long-horizon settings where proprioceptive visual grounding is most useful. Across both CALVIN and LIBERO-Long, the retained patches move with the task phase: they cover task-referenced objects and target regions during approach, then shift toward the gripper, drawer, and contact regions during manipulation. These examples complement the aggregate long-horizon metrics by showing that token retention remains state-conditioned over multi-stage execution rather than staying fixed on the initial target object.

\section{Analysis and Ablations}
\label{sec:exp-analysis}

\subsection{Inference-Time Breakdown}

\begin{table}[ht]
  \centering
  \caption{Mean per-environment-step latency in ms on CALVIN and LIBERO. Tokens reports the average retained visual-token count over the total available visual tokens.}
  \label{tab:latency_breakdown}
  \small
  \setlength{\tabcolsep}{5pt}
  \begin{tabular}{@{}lcccccc@{}}
    \toprule
    \textbf{Benchmark} & \textbf{Tokens} & \textbf{Vision} & \textbf{Selector} & \textbf{VLM} & \textbf{Action} & \textbf{Total} \\
    \midrule
    CALVIN ABC$\to$D & 12/100 & 5.8 & 0.1 & 0.9 & 15.4 & 22.2 \\
    LIBERO & 7.8/34 & 6.5 & 0.1 & 1.1 & 16.1 & 23.8 \\
    \bottomrule
  \end{tabular}
\end{table}

\Cref{tab:latency_breakdown} decomposes the end-to-end inference time used in \Cref{tab:efficiency}. The token denominators reflect the two-view preprocessing resolutions: CALVIN uses $224\times224$ inputs and produces 100 visual tokens per timestep, whereas LIBERO uses $112\times112$ inputs and produces 34. The Vision column includes encoding both camera views. The selector adds only about $0.1$ ms per step in both benchmarks, while the diffusion action head remains the dominant cost. The total latency is therefore reduced mainly by shortening the sequence processed by the VLM and consumed by the cross-attention action head, rather than by changing the action generator itself.

\subsection{Proprioceptive Discretization Sensitivity}
\label{app:bin-clip-sensitivity}

\begin{table}[ht]
  \centering
  \caption{Auxiliary LIBERO-Spatial sensitivity runs for proprioceptive discretization. We vary the number of bins and clipping range used before mapping state values to VLM-vocabulary token IDs.}
  \label{tab:bin-clip-sensitivity}
  \small
  \setlength{\tabcolsep}{8pt}
  \begin{tabular}{@{}ccc@{}}
    \toprule
    \textbf{Bins} & \textbf{Clip range} & \textbf{Success rate (\%)} $\uparrow$ \\
    \midrule
    32  & $[-3,3]$ & $94.5{\pm}0.1$ \\
    256 & $[-3,3]$ & $98.4{\pm}0.2$ \\
    512 & $[-3,3]$ & $98.4{\pm}0.3$ \\
    256 & $[-1,1]$ & $96.1{\pm}0.9$ \\
    256 & $[-5,5]$ & $96.2{\pm}0.6$ \\
    \bottomrule
  \end{tabular}
\end{table}

\Cref{tab:bin-clip-sensitivity} shows that proprioceptive tokenization is sensitive to both quantization resolution and clipping range. Too few bins coarsen the state representation, and too narrow or too wide clipping ranges reduce useful variation in the resulting token IDs. Increasing from 256 to 512 bins does not improve performance in this setup, so we use 256 bins over $[-3,3]$ as the default. These are the defaults used by the VLM-vocabulary proprioceptive encoding in \cref{sec:method-encoding} and by the observation/model setup in \cref{app:observations-proprio,app:model-architecture}. This supports the view that VLM-vocabulary proprioception is not simply a free replacement for continuous state: the discretization must retain enough resolution for the robot state while remaining stable under the normalized proprioceptive distribution.

\subsection{Selection Consistency}

\begin{table}[t]
  \centering
  \caption{Intersection-over-Union (IoU) between binary retained-token masks. Cross-scene compares the same instruction across different initial states; Temporal compares neighboring rollout steps; Random is the expected IoU from uniform random retention with the matched token budget (12/100 for CALVIN ABC$\to$D, 7.8/34 for LIBERO-Spatial).}
  \label{tab:token-consistency}
  \small
  \setlength{\tabcolsep}{6pt}
  \begin{tabular}{@{}lcc@{}}
    \toprule
    \textbf{IoU} & \textbf{\begin{tabular}{@{}c@{}}CALVIN\\ABC$\to$D\end{tabular}} & \textbf{\begin{tabular}{@{}c@{}}LIBERO\\Spatial\end{tabular}} \\
    \midrule
    Cross-scene & 0.170 & 0.644 \\
    Temporal & 0.196 & 0.418 \\
    Random & 0.064 & 0.130 \\
    \bottomrule
  \end{tabular}
\end{table}

\Cref{tab:token-consistency} quantifies whether retained-token masks are structured rather than arbitrary. Each mask records the visual tokens kept in one forward pass, so higher IoU means greater overlap between selected token sets; the matched token budgets are 12/100 for CALVIN ABC$\to$D and 7.8/34 for LIBERO-Spatial. Cross-scene and temporal IoU are both above the random baseline on CALVIN and LIBERO-Spatial, indicating that selection is task- and state-conditioned. The lower CALVIN cross-scene IoU reflects stronger variation across long-horizon configurations, while the higher LIBERO-Spatial IoU is consistent with the more constrained scene layouts in that suite.

\subsection{Phase-Wise Selection Adaptation}

\begin{figure}[t]
  \centering
  \includegraphics[width=0.68\columnwidth]{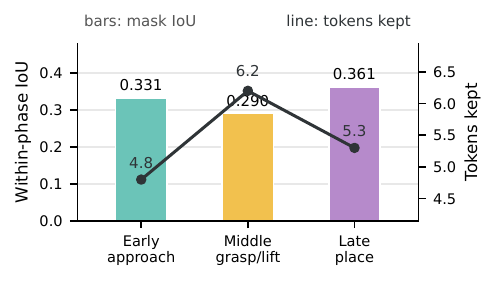}
  \caption{Phase-wise selector behavior on LIBERO-Spatial. Bars show within-phase mask IoU, and the line shows mean tokens kept. Phases are equal thirds of each rollout.}
  \label{fig:phasewise-selection}
\end{figure}

\Cref{fig:phasewise-selection} shows that the selector is not static over an episode. The mean-token line peaks in the middle phase, and the middle-phase IoU bar is the lowest, consistent with grasping and lifting being the most visually dynamic portion of the task. Late-phase masks become more consistent as the robot moves toward the placement target, and the Early--Late IoU bar is approximately 0.29, indicating substantial cross-phase drift. This phase-wise view complements the qualitative rollouts in \cref{fig:calvin-lh5-rollout,fig:libero-long-rollout}, where retained tokens shift with objects, gripper motion, and contact regions over execution.

\subsection{Qualitative Failure Cases}
\label{app:qualitative-failure-cases}

\begin{figure}[t]
  \centering
  \includegraphics[width=0.92\columnwidth]{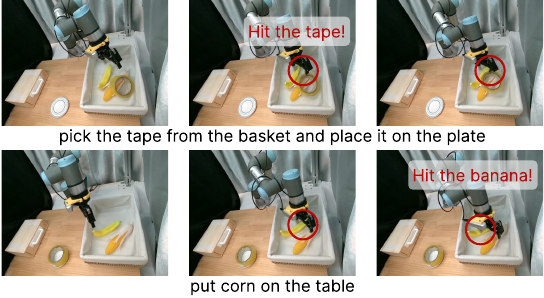}
  \caption{Representative real-world failure cases. Top: the gripper reaches the tape but makes misaligned contact. Bottom: the approach to the corn collides with a nearby banana.}
  \label{fig:fail-cases}
\end{figure}

\Cref{fig:fail-cases} shows two representative real-world failures. In the tape case, the gripper reaches the correct object but contacts it with a misaligned approach, so the final manipulation fails despite the target being visually identified. In the corn case, the approach trajectory clips a nearby banana before establishing clean contact with the corn. These examples point to control-side and contact-side limitations, complementing the main-paper \hyperref[sec:limitations]{Limitations section}, which focuses on the tabletop evaluation scope and deployment assumptions.

\end{document}